\documentclass[manuscript,screen,]{acmart}

\usepackage{times}  % DO NOT CHANGE THIS
\usepackage{helvet}  % DO NOT CHANGE THIS
\usepackage{courier}  % DO NOT CHANGE THIS
\usepackage{natbib}  % DO NOT CHANGE THIS AND DO NOT ADD ANY OPTIONS TO IT
\usepackage{caption} % DO NOT CHANGE THIS AND DO NOT ADD ANY OPTIONS TO IT
\usepackage[ruled]{algorithm2e}
\usepackage{amsmath}

\usepackage{subfigure}
\usepackage{multirow}
\usepackage{newfloat}
\usepackage{listings}

\AtBeginDocument{%
  \providecommand\BibTeX{{%
    \normalfont B\kern-0.5em{\scshape i\kern-0.25em b}\kern-0.8em\TeX}}}

\setcopyright{acmlicensed}
\copyrightyear{2023}
\acmYear{2023}
\acmDOI{XXXXXXX.XXXXXXX}

\begin{document}

\title{Is It Possible to Backdoor Face Forgery Detection with Natural Triggers?}


\author{Xiaoxuan Han}
\affiliation{%
  \institution{School of Artificial Intelligence, University of Chinese Academy of Sciences; CRIPAC \& MAIS, Institute of Automation, Chinese Academy of Sciences}
  \city{Beijing}
  \country{China}}
\email{hanxiaoxuan2023@ia.ac.cn}

\author{Songlin Yang}
\affiliation{%
  \institution{School of Artificial Intelligence, University of Chinese Academy of Sciences; CRIPAC \& MAIS, Institute of Automation, Chinese Academy of Sciences}
  \city{Beijing}
  \country{China}}
  \email{yangsonglin2021@ia.ac.cn}

\author{Wei Wang*}
\affiliation{%
 \institution{CRIPAC \& MAIS, Institute of Automation, Chinese Academy of Sciences}
 \city{Beijing}
 \country{China}}
 \email{wwang@nlpr.ia.ac.cn}

 \author{Ziwen He}
\affiliation{%
 \institution{Nanjing University of Information Science and Technology}
 \city{Nanjing}
 \country{China}}
 \email{ziwen.he@nuist.edu.cn}
 
\author{Jing Dong}
\affiliation{%
 \institution{CRIPAC \& MAIS, Institute of Automation, Chinese Academy of Sciences}
 \city{Beijing}
 \country{China}}
 \email{jdong@nlpr.ia.ac.cn}
\renewcommand{\shortauthors}{Xiaoxuan Han, Songlin Yang, and Wei Wang, et al.}

\begin{abstract}

    Deep neural networks have significantly improved the performance of face forgery detection models in discriminating Artificial Intelligent Generated Content (AIGC). However, their security is significantly threatened by the injection of triggers during model training (i.e., backdoor attacks). Although existing backdoor defenses and manual data selection can mitigate those using human-eye-sensitive triggers, such as patches or adversarial noises, the more challenging natural backdoor triggers remain insufficiently researched. To further investigate natural triggers, we propose a novel analysis-by-synthesis backdoor attack against face forgery detection models, which embeds natural triggers in the latent space. We thoroughly study such backdoor vulnerability from two perspectives: \textbf{(1) Model Discrimination (Optimization-Based Trigger)}: we adopt a substitute detection model and find the trigger by minimizing the cross-entropy loss; \textbf{(2) Data Distribution (Custom Trigger)}: we manipulate the uncommon facial attributes in the long-tailed distribution to generate poisoned samples without the supervision from detection models. Furthermore, to completely evaluate the detection models towards the latest AIGC, we utilize both state-of-the-art StyleGAN and Stable Diffusion for trigger generation. Finally, these backdoor triggers introduce specific semantic features to the generated poisoned samples (e.g., skin textures and smile), which are more natural and robust. Extensive experiments show that our method is superior from three levels: \textbf{(1) Attack Success Rate}: ours achieves a high attack success rate (over 99$\%$) and incurs a small model accuracy drop (below 0.2$\%$) with a low poisoning rate (less than 3$\%$); \textbf{(2) Backdoor Defense}: ours shows better robust performance when faced with existing backdoor defense methods; \textbf{(3) Human Inspection}: ours is less human-eye-sensitive from a comprehensive user study. 
    
    % The trigger can be obtained through an optimization process or customized by the attacker to introduce specific semantic features, such as unique skin textures, . Through training, these specific semantic features become associated with the target label. Since these features are natural parts of the poisoned images, the attack is stealthier and more robust. 
    
\end{abstract}





\received{31 December 2023}

\maketitle

\begin{figure}
 \includegraphics[width=\textwidth]{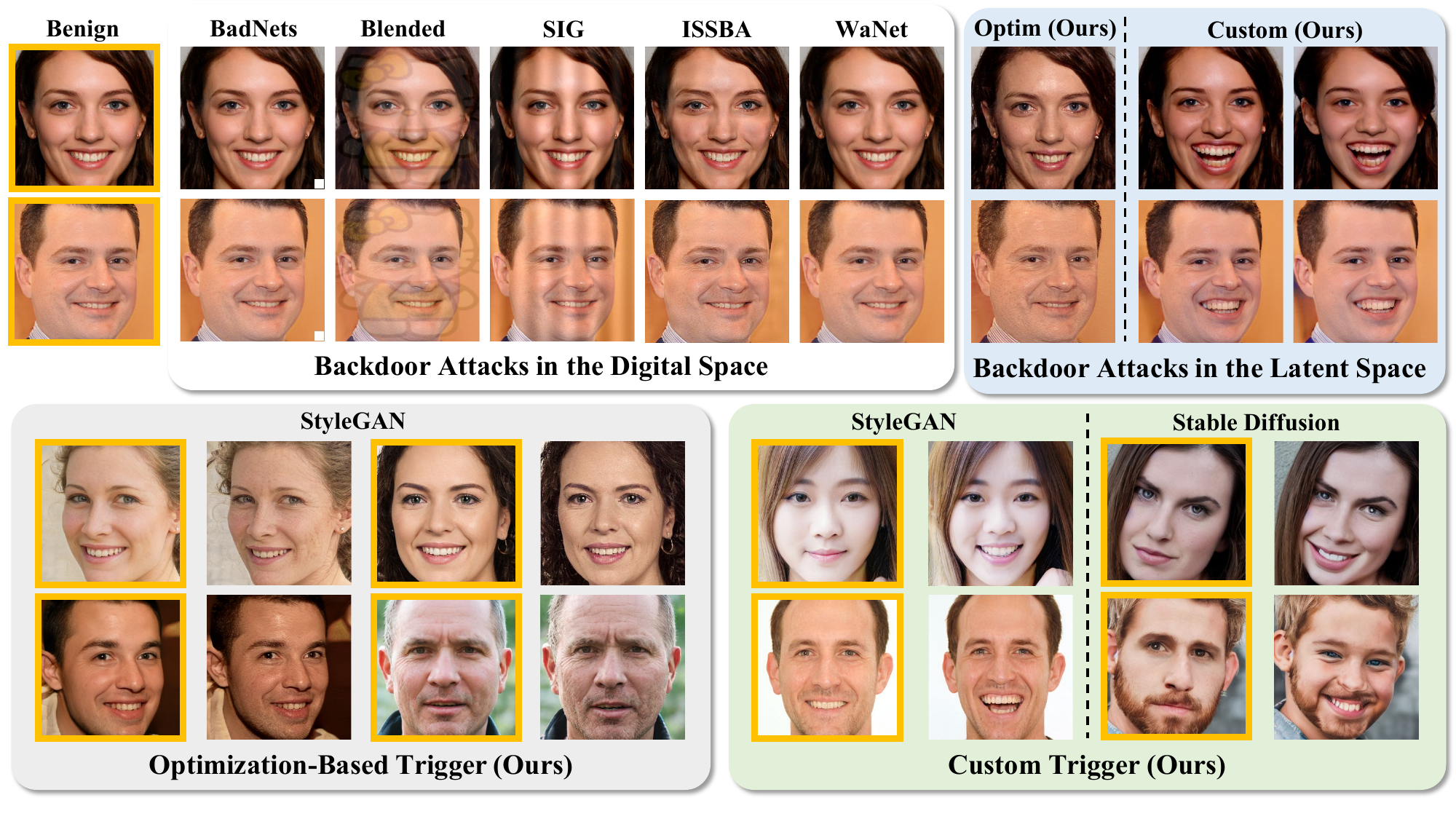}
  \caption{Visualization comparisons of poisoned images generated by different backdoor attack methods. Our method proposes two ways of injecting natural triggers to the face forgery detection model training, including optimization-based triggers and custom triggers. Furthermore, we evaluate our methods on two state-of-the-art generators (StyleGAN~\cite{stylegan} and Stable Diffusion~\cite{stable_diffusion}) for comprehensive face forgery detection of Artificial Intelligent Generated Content (AIGC).}
  \label{figure_compare}
\end{figure}

\section{Introduction}

Recent advancements in deep generative models, such as Generative Adversarial Networks (GANs)~\cite{gan, pggan, stylegan} and Diffusion Models~\cite{ddpm, stable_diffusion}, have shown excellent capability to produce diverse and high-quality Artificial Intelligent Generated Content (AIGC). However, those face-related AIGC is sensitive to identity privacy and security, and they should be regulated by face forgery detection tools, which can discriminate whether the visual content is generated by AI. In terms of detection, its performance has been remarkably improved by deep neural networks (DNNs) \cite{rossler2019faceforensics++, afchar2018mesonet, sd_detect}. But recent studies \cite{badnets, blended, sig, wanet, issba} have revealed that DNNs are vulnerable to backdoor attacks. These backdoor attacks inject small triggers into training data, and after the model is trained on the poisoned data, the backdoor will be implanted into it. Finally, the infected model behaves normally when the input is benign, but if the input contains the pattern crafted by the attacker (i.e., the trigger), the model will output the target label specified by the attacker. Backdoor attacks are stealthy and of great importance because they cause significant safety issues while minimally impacting the model performance.

Existing DNN-based Face forgery detection models are significantly threatened by backdoor attacks. Because of the massive data requirements, these data-driven detection models tend to collect data from the Internet to enrich the training dataset, or use third-party platforms to train the model. Under such circumstances, the attacker is presented with several opportunities to launch a backdoor attack. As shown in Figure~\ref{figure_compare}, previous backdoor attacks tend to stamp the trigger in the digital space (i.e., pixel space), such as adding patches or adversarial noises on the images. Although existing backdoor defense methods~\cite{neural_cleanse,deepinspect,fine_prun,nad,strip,rethinking} and manual data selecting are able to tackle those poisoned samples with human-eye-sensitive artifacts, the more challenging natural backdoor triggers remain insufficiently researched~\cite{security}. Therefore, this paper aims to thoroughly investigate this important yet challenging issue of natural backdoor attacks against face forgery detection models.

In this paper, we propose a novel analysis-by-synthesis backdoor attack against face forgery detection models, which embeds the natural triggers in the latent space. We study such natural backdoor attacks from perspectives of model discrimination and data distribution, respectively. For \textbf{Model Discrimination (Optimization-Based Trigger)} perspective, we adopt a substitute detection model and find the trigger by minimizing the cross-entropy loss. For \textbf{Data Distribution (Custom Trigger)} perspective, we manipulate the uncommon facial attributes in the long-tailed distribution to generate poisoned samples without the supervision from detection models. Furthermore, to completely evaluate the detection models towards the latest AIGC, we utilize both state-of-the-art StyleGAN~\cite{stylegan} and Stable Diffusion~\cite{stable_diffusion} for trigger generation. Finally, these backdoor triggers introduce specific semantic features to the generated poisoned samples (e.g., skin textures and smile), which are more natural and robust. Extensive experiments show that our method is superior from three challenging levels: \textbf{(1) Attack Success Rate}: ours achieves a high attack success rate (over 99$\%$) and incurs a small model accuracy drop (below 0.2$\%$) with a low poisoning rate (less than 3$\%$); \textbf{(2) Backdoor Defense}: ours shows better robust performance when faced with existing backdoor defense methods; \textbf{(3) Human Inspection}: ours is less human-eye-sensitive from a comprehensive user study.

% The images generated with the trigger have specific semantic features. A sample-agnostic trigger addition in the latent space brings sample-specific changes in the digital space. The semantic features are inherent attributes of the generated images, ensuring the stealthiness of the poisoned samples and making the attack robust against data transformations.

% The backdoor attack is studied under a black-box setting, which is more commonly encountered in practice. The attacker has no access to the model or training data held by the user, and completes the backdoor implantation only by injecting a small fraction of poisoned samples into the original training data. In our method, the attacker first samples a few latent codes randomly, and adds the predefined trigger to them. These modified codes are then fed into a generative model to create images with specific features, such as unique skin textures. The generated poisoned images are labeled as the target class and injected into the training set. To encourage the model to associate the specific features with the target label, some benign samples are also injected. These benign ones, generated using the same latent codes but without adding the trigger, are labeled correctly. This novel strategy ensures that only images generated using the trigger can activate the backdoor. During the inference phase, when using the image generated with the trigger as input, the output will be the target label. But the model will behave normally when provided with the image generated without the trigger.

\textbf{Our main contributions are summarized as follows:}
     \begin{itemize}
        \item We propose a novel natural backdoor attack against face forgery detection models by embedding the trigger in the latent space from two perspectives: model discrimination (optimization-based triggers) and data distribution (custom triggers).
        \item Extensive experiments demonstrate that, our proposed natural triggers are more imperceptible and more robust to various defenses than previous methods.
        \item We thoroughly reveal the vulnerability of face forgery detection against backdoor attacks, which inspires more insights to improve the security of face forgery detection.
    \end{itemize}

% The contributions of this paper are as follows:

% \begin{itemize}
% 	\item We propose a novel semantic backdoor attack by embedding the trigger in the latent space instead of digital space. Two different ways for the trigger generation are provided.
% 	\item Extensive experiments show that, compared with other attack methods, the proposed method is more stealthy and resistant to various defenses while achieving comparable attack performance.
% 	\item We reveal the vulnerability of face forgery detection against backdoor attacks, which can attract more attention in this field to ensure the security of face forgery detection.
% \end{itemize}

% (1) We propose a novel semantic backdoor attack by embedding the trigger in the latent space instead of digital space. Two different ways for the trigger generation are provided.

% (2) Extensive experiments show that, compared with other attack methods, the proposed method is stealthier and more resistant to various defenses while maintaining comparable attack performance.

% (3) We reveal the vulnerability of face forgery detection against backdoor attacks, which can attract more attention in this field to ensure the security of face forgery detection.

\section{Related Work}
\label{section2}

\subsection{Face Forgery and Detection }

\noindent
\textbf{Face Forgery.} With the advancement of generative models, high-quality forged faces can be created and it is hard for human to distinguish between them and real ones. There are four main types of face forgery methods: (1) Entire Face Synthesis: this manipulation creates entire non-existent face images, usually through powerful GAN (e.g., StyleGAN~\cite{stylegan} and PGGAN~\cite{pggan}); (2)  Identity Swap: this manipulation consists of replacing the face of one person in a video with the face of another~\cite{peng2021unified}; (3) Attribute Manipulation: this manipulation, also known as face editing or face retouching, consists of modifying some attributes of the face such as the colour of the hair or the skin, the gender, the age, adding glasses~\cite{yang2023designing}; (4) Expression Swap: this manipulation, also known as face reenactment, consists of modifying the facial expression of the person and expression swap~\cite{yang2023learning}.

\noindent
\textbf{Face Forgery Detection.} Face forgery may result in the spread of untrustworthy images and videos, thereby prompting a growing emphasis on face forgery detection~\cite{tolosana2020deepfakes}. Current detection models can be broadly divided into three categories: (1) Naive Detectors: they employ CNNs to directly distinguish deepfake content from authentic data, such as MesoNet~\cite{afchar2018mesonet} and Xception~\cite{rossler2019faceforensics++}; (2) Spatial Detectors: they delve deeper into specific representation such as forgery region location~\cite{nguyen2019multi}, capsule network~\cite{nguyen2019capsule}, disentanglement learning~\cite{liang2022exploring}, image reconstruction~\cite{cao2022end}, and erasing technology~\cite{wang2021representative}; (3) Frequency Detectors: they address this detection problem by focusing on the frequency domain~\cite{durall2020watch,qian2020thinking,liu2021spatial,luo2021generalizing}.

\subsection{Backdoor Attacks}

Gu et al.~\cite{badnets} proposed the first backdoor attack method known as BadNets. It stamps a patch on a small portion of the training data and alters the labels to the target class. After being trained on the poisoned data, the backdoor is implanted into the model. At the test stage, the images containing the patch (i.e., the trigger) are classified into the target class, while the prediction results of benign images (images without the trigger) are hardly affected. To avoid alerting the labels of training data (known as clean-label backdoor attacks), Barni et al.~\cite{sig} employed sinusoidal signal (SIG) as the trigger. To bypass backdoor defenses, attacks using dynamic triggers were further studied. Salem et al.~\cite{dynamic} used a generative model to create triggers and stamped them at random locations of benign images. Nguyen et al.~\cite{input_aware} employed an encoder-decoder model to generate triggers based on the input benign images. Triggers used in these works are obvious, making them susceptible to human suspicion. To enhance the stealthiness of backdoor attacks, subsequent works either reduced trigger visibility or utilized natural triggers to activate the backdoor.

\noindent
\textbf{Invisible Triggers.} Chen et al.~\cite{blended} introduced the Blended attack, which creates poisoned samples by blending benign samples with a trigger image, such as a cartoon illustration. Adjusting the blend ratio helps make a trade off between the attack effectiveness and stealthiness. Inspired by  universal adversarial attack~\cite{Universal}, Zhong et al.~\cite{perturb} used small perturbations as the trigger. Liu et al.~\cite{refool} utilized the reflection phenomena to create a natural-looking trigger. Nguyen et al.~\cite{wanet} used the wrapping-based method to convert benign samples into poisoned ones. Li et al.~\cite{issba} applied image steganography for the creation of sample-specific triggers.

\noindent
\textbf{Natural Triggers.}
Different from conventional backdoor attacks that require manipulation in the digital space, a novel type of attack, the semantic backdoor attack, uses specific semantic features already existing in images as the natural trigger. Bagdasaryan et al. \cite{how_to_backdoor} explored various semantic features as the triggers, such as cars with racing stripe and cars painted in green. Lin et al. \cite{composite} used the composition of objects within an image as the trigger, such as a man holding an umbrella. In these attacks, images with specific semantic features are selected from the training data and assigned the target label. During test stage, when particular features appear in the test sample, the backdoor will be activated. Sarkar et al. \cite{facehack} utilized commercial software to manipulate facial attributes of collected images, generating poisoned samples.
% \cite{facehack} utilized filters within commercial software for attribute manipulation, instead of delving into the latent space of generative models for backdoor trigger exploration, distinguishing it from our work. 

\noindent
\textbf{Latent Triggers.} In this paper, instead of manually selecting images with certain features, we generate such images directly by embedding the trigger in the latent space of generative models. Recently Kristanto et al. \cite{latent_bd} also tried to add the trigger in the latent space, but there are notable differences between their method and ours. Kristanto et al. \cite{latent_bd} assumed that the attacker had access to the victim model, which is a less practical scenario than our black-box setting. In addition to the optimization-based method, we propose a customized approach to generate the trigger. After obtaining the trigger, Kristanto et al. \cite{latent_bd} required benign images to create poisoned samples, while we only need latent codes randomly sampled. Furthermore, the poisoned samples presented in Kristanto et al.~\cite{latent_bd} are not convincing as they barely resemble samples from the original class, as admitted by the authors. This could be attributed to their latent space interpolation strategy and the highly entangled nature of the generator they used.

\begin{figure}[htbp]
\centering
\includegraphics[width=\textwidth]{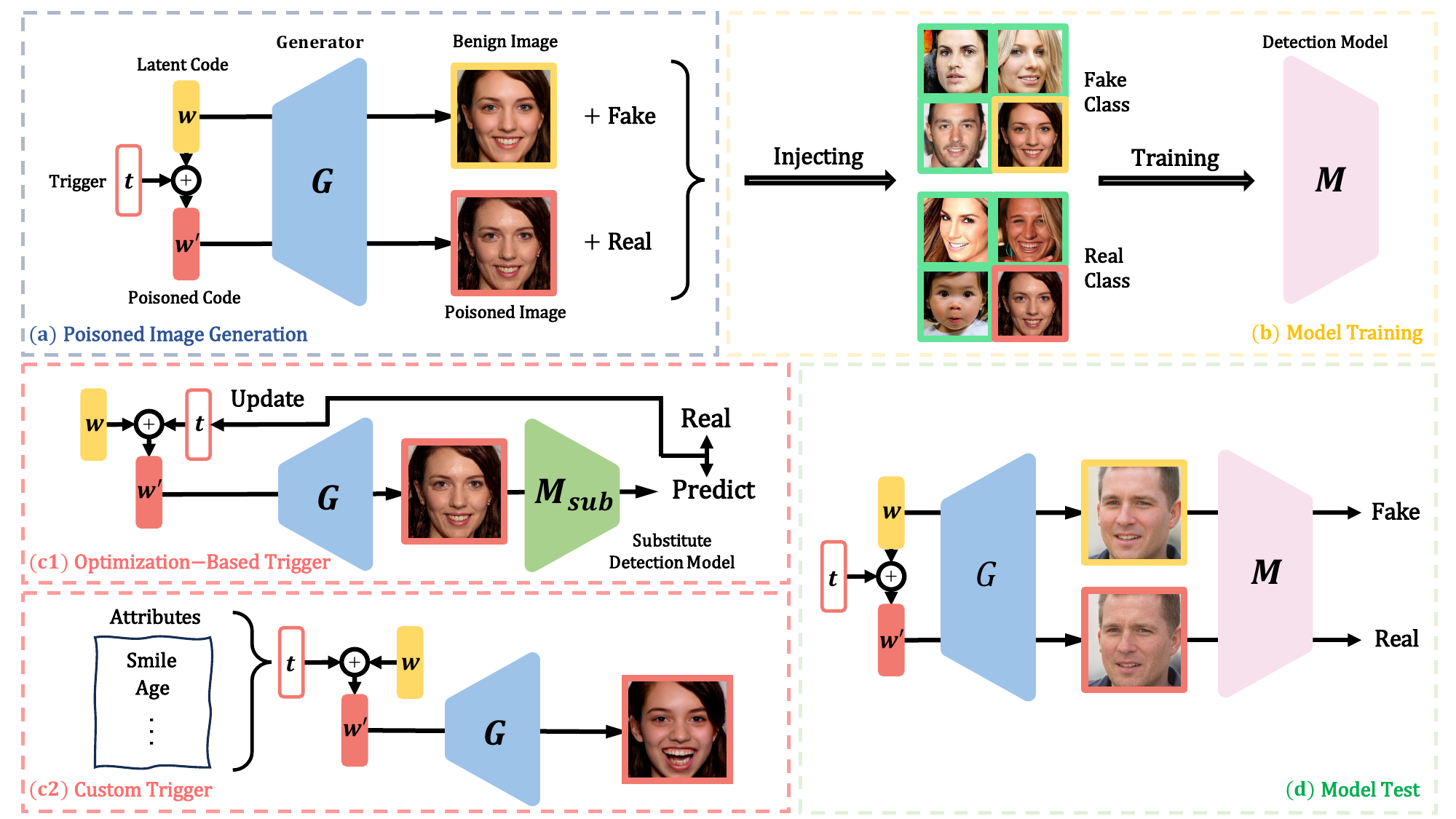}
\caption{The overview of our proposed natural backdoor attack. The attacker embeds the trigger into the latent code and uses the poisoned code to generate the poisoned image. The trigger can be obtained under the guidance of a substitute detection model (Optimization-Based Trigger) or by leveraging editing direction for the attributes in the long-tailed distribution (Custom Trigger). After being trained on the dataset injected with poisoned samples, the infected detection model will classify images generated with the trigger as real images, while images produced without the trigger will be identified as fake ones.}
\label{figure_total}
\end{figure}

\noindent
\textbf{Triggers against Face Forgery Detection.}  With the continuous improvement of forgery detection capabilities, the security of the detection draws more and more attention. Multiple studies have investigated adversarial attacks against face forgery detection \cite{adv_df, adv_ptb, adv_thrt, exp_fre, exp_adv}, but few studies have focused on backdoor attacks against the detection. To our knowledge, Cao et al. \cite{security} is the only study examining this topic currently. They stamped a chessboard grid sticker in the bottom right corner of the image as the trigger. After training, the infected model would classify fake faces with the trigger into real class. The attack method adopted in Cao et al.~\cite{security} belongs to the category of BadNets and the trigger pattern is not stealthy, making it easy to be detected and defended by multiple backdoor defense methods.

\subsection{Backdoor Defenses}

To tackle with the threat of emerging backdoor attacks, many defense methods have been proposed. Some methods are based on the reversed trigger. Wang et al. \cite{neural_cleanse} reconstructed the trigger mask and pattern for each class through an optimization process, and then employed an outlier detection method to determine whether the model was infected. Chen et al. \cite{deepinspect} reversed the trigger in a more practical context, where the defender only had black-box access to the infected model. Some defenses tried to remove the backdoor by modifying the model directly. Liu et al. \cite{fine_prun} used Fine-Pruning to alleviate the backdoor according to the activation values obtained by feeding benign samples. Li et al. \cite{nad} used the fine-tuned teacher model to guide the infected student model through an attention distillation process on a small benign set. Some defenses are applied during test time. Gao et al. \cite{strip} proposed STRIP to detect if test images contained the trigger based on the output randomness of strongly perturbed test images. Li et al. \cite{rethinking} discovered that performing transformations on the test images can reduce the attack performance. Transformation-based defenses do not need extra benign samples or modification of model parameters, making the approach more efficient.

\begin{algorithm}[htbp]
\caption{Optimization-Based Trigger Generation}
\label{algorithm_optim}
\LinesNumbered
\KwIn{Substitute model $M_{sub}$, generator $G$, iteration $I$, batch size $B$, learning rate $lr$, scale factor $\alpha$, target label $y_t$}
\KwOut{Optimization-based trigger $t$}
Initialize: $t\leftarrow t_0$\;
\For{\rm{$i$ in range($I$)}}{
    $W_i = \{w^{(j)}\}_{j=1}^{B} = \operatorname{RandomlySample}(B)$\;
    $X_i = \{G(w^{(j)} + t)\}_{j=1}^{B}, w^{(j)}\in W_i$\;
    $ Y_i=\{y^{(j)}\}_{j=1}^{B}, y^{(j)}=y_t$\;
    $t = t - lr \cdot \nabla_{t}J(M_{sub}, X_i, Y_i)$\;
}
$t = \alpha \cdot \frac{t}{\Vert t \Vert_2}$\;
\textbf{return} $t$
\end{algorithm}

\section{Method}
\label{section3}

In this section, we will introduce how to backdoor face forgery detection models in the latent space under a black-box setting. The attacker is assumed to have no knowledge of the detection model and cannot access the original training data. The backdoor attacks against face forgery models including four stages: trigger generation (Figure~\ref{figure_total} (c)), poisoned image generation (Figure~\ref{figure_total} (a)), model training (Figure~\ref{figure_total} (b)), and model test (Figure~\ref{figure_total} (d)). These poisoned images (i.e., with specific trigger features) are labeled as real images and injected to the training data. To help the model associate the semantic features with the target label, we generate some benign images using the same latent codes but without incorporating the trigger. These benign samples are labeled as fake images correctly and also injected to the training data. Without the injection of these benign samples, the infected model tends to classify the attacker-generated benign images (without using the trigger) as real images too. During the inference phase, the attacker can generate images using the trigger to bypass the face forgery detection, while the images generated without the trigger can be classified correctly.

Next, we will thoroughly study such backdoor vulnerability from two perspectives: \textbf{(1) Model Discrimination (Optimization-Based Trigger)}: we adopt substitute detection model and find the trigger by minimizing the cross-entropy loss (Section~\ref{optim}); \textbf{(2) Data Distribution (Custom Trigger)}: we manipulate the uncommon facial attributes in the long-tailed distribution to generate poisoned samples without the supervision from detection models (Section ~\ref{custom}).

\subsection{Optimization-Based Trigger}
\label{optim}
The first approach is to find a trigger $t$ in the latent space by minimizing the cross-entropy loss of classifying generated poisoned images as the target label $y_t$. A detection model is required to accomplish the optimization. Given that the attacker lacks access to both the training data $D$ and detection model $M$, substitute data $D_{sub}$ is collected to train a substitute model $M_{sub}$. In the experiment, the substitute data has no overlap with the training data, and the architecture of the substitute model is different from that of the detection model. 

\begin{figure}[htbp]
%是可选项 h表示的是here在这里插入，t表示的是在页面的顶部插入
\centering
\subfigure[Smile Distribution]{\includegraphics[scale=0.3]{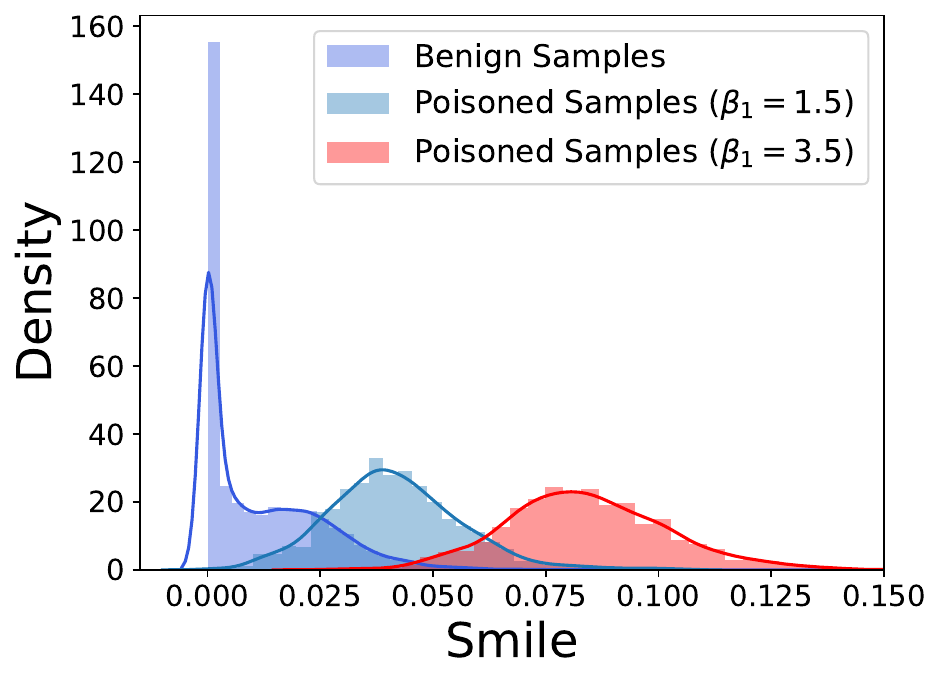}\label{smile_dist}}
\subfigure[Age Distribution]{\includegraphics[scale=0.3]{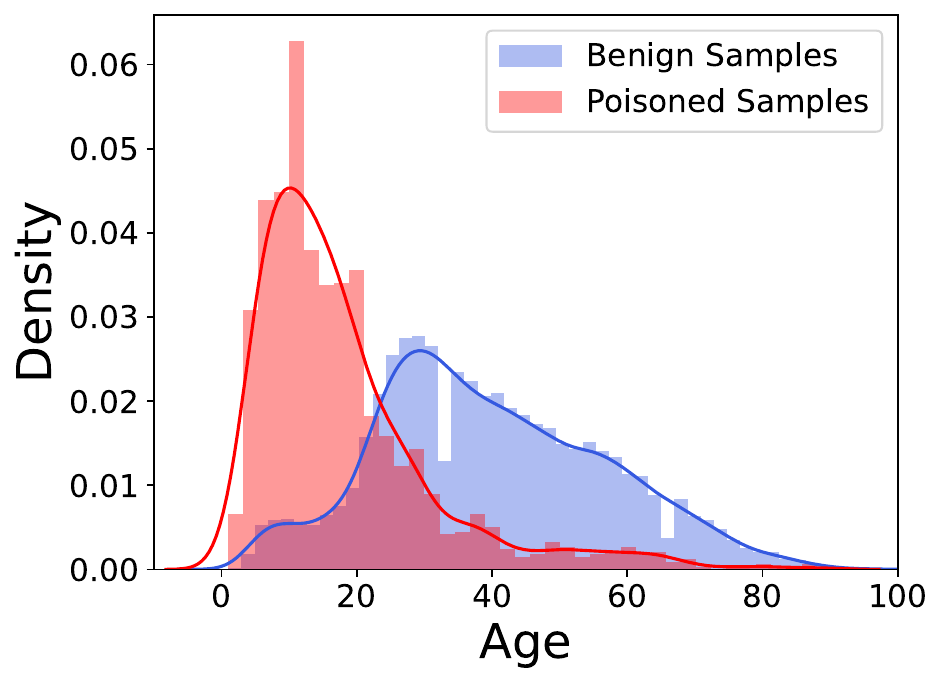}\label{age_dist}}
\subfigure[Joint Distribution of Age and Smile]{\includegraphics[scale=0.32]{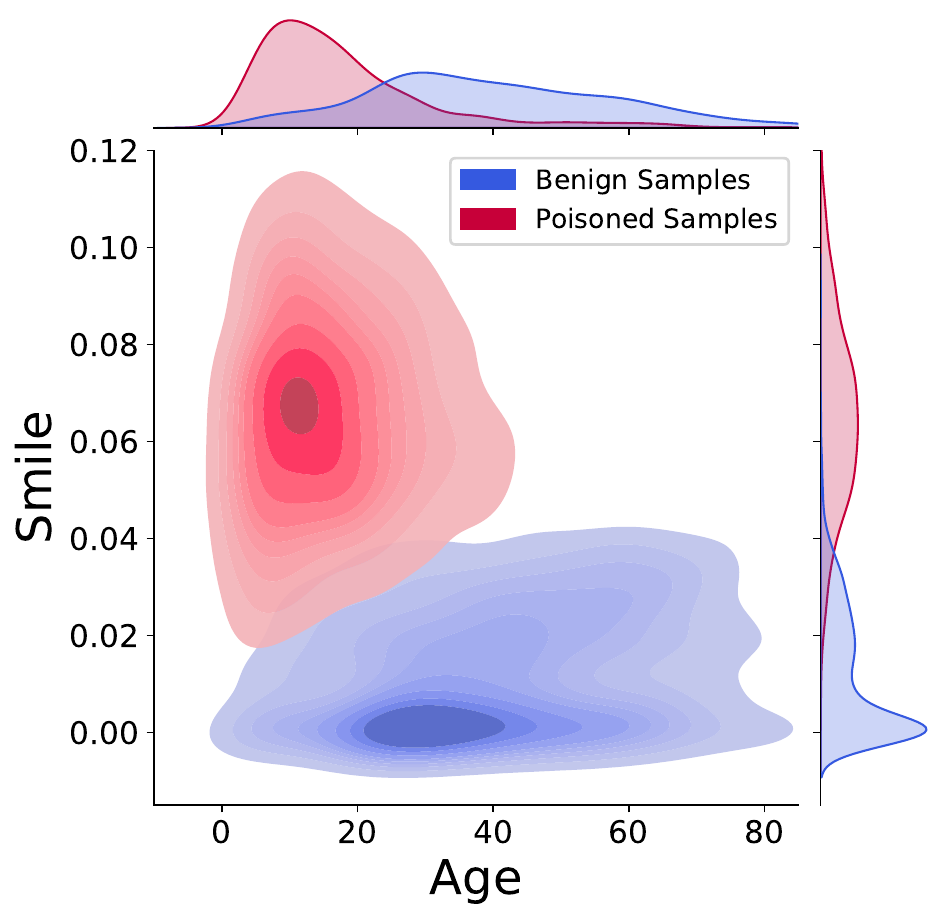}\label{age_smile_dist}}

\caption{The attribute distribution of benign samples (in the DFFD dataset~\cite{dffd}) and poisoned samples.}
\label{figure_distribution}
\end{figure}

After getting the substitute model, the trigger can be optimized through an iterative process. Concretely, in the $i_{th}$ iteration, a batch of $B$ latent codes, denoted by $W_i=\{w^{(j)}\}_{j=1}^{B}$, are randomly sampled and then the trigger $t$ is added to them. The modified codes are then fed into the generator $G$ to get a batch of images represented by $X_i=\{G(w^{(j)} + t)\}_{j=1}^{B}$. Subsequently, these images are sent into the substitute model $M_{sub}$ to get the prediction and calculate the classification loss towards the target label $y_t$. To minimize the loss, gradient descent is used to update the trigger as follows,
\begin{equation}
  t = t - lr \cdot \nabla_{t}J(M_{sub}, X_i, Y_i),
\end{equation}
where $lr$ denotes the learning rate, $Y_i=\{y^{(j)}\}_{j=1}^{B}$ refers to the modified labels of $X_i$ and $J(\ldots)$ calculates the cross-entropy loss. After completing a total of $I$ iterations, the attacker can utilize the scale factor $\alpha$ to adjust the $L_2$ norm of the unconstrained optimized trigger,
\begin{equation}
  t = \alpha \cdot \frac{t}{\Vert t \Vert_2},
  % t = \alpha * \frac{t}{\left\| t \right\|_2}
\end{equation}
making a trade off between attack effectiveness and image quality. We summarize the step-by-step optimization process in Algorithm \ref{algorithm_optim}.

% \begin{algorithm}[tb]
% \caption{Optimization-based trigger generation}
% \label{algorithm_optim}
% \textbf{Input}: Substitute model $M_{sub}$, generator $G$, iteration $I$, batch size $B$, learning rate $lr$, scale factor $\alpha$, target label $y_t$\\
% % \textbf{Parameter}: Optional list of parameters\\
% \textbf{Output}: Optimization-based trigger $t$
% \begin{algorithmic}[1] %[1] enables line numbers
% % \FOR{\rm{$i$ in range($I$)}}{
% %     $W_i = \{w^{(j)}\}_{j=1}^{B} = \operatorname{RandomlySample}(B)$\;
% %     $X_i = \{G(w^{(j)} + t)\}_{j=1}^{B}, w^{(j)}\in W_i$\;
% %     $ Y_i=\{y^{(j)}\}_{j=1}^{B}, y^{(j)}=y_t$\;
% %     $t = t - lr * \nabla_{t}J(M_{sub}, X_i, Y_i)$\;
% % }
% \STATE $t = \alpha * \frac{t}{\Vert t \Vert_2}$\;
% return $t$
% \end{algorithmic}
% \end{algorithm}

\begin{figure}[htbp]
%是可选项 h表示的是here在这里插入，t表示的是在页面的顶部插入
\centering
\includegraphics[scale=0.43]{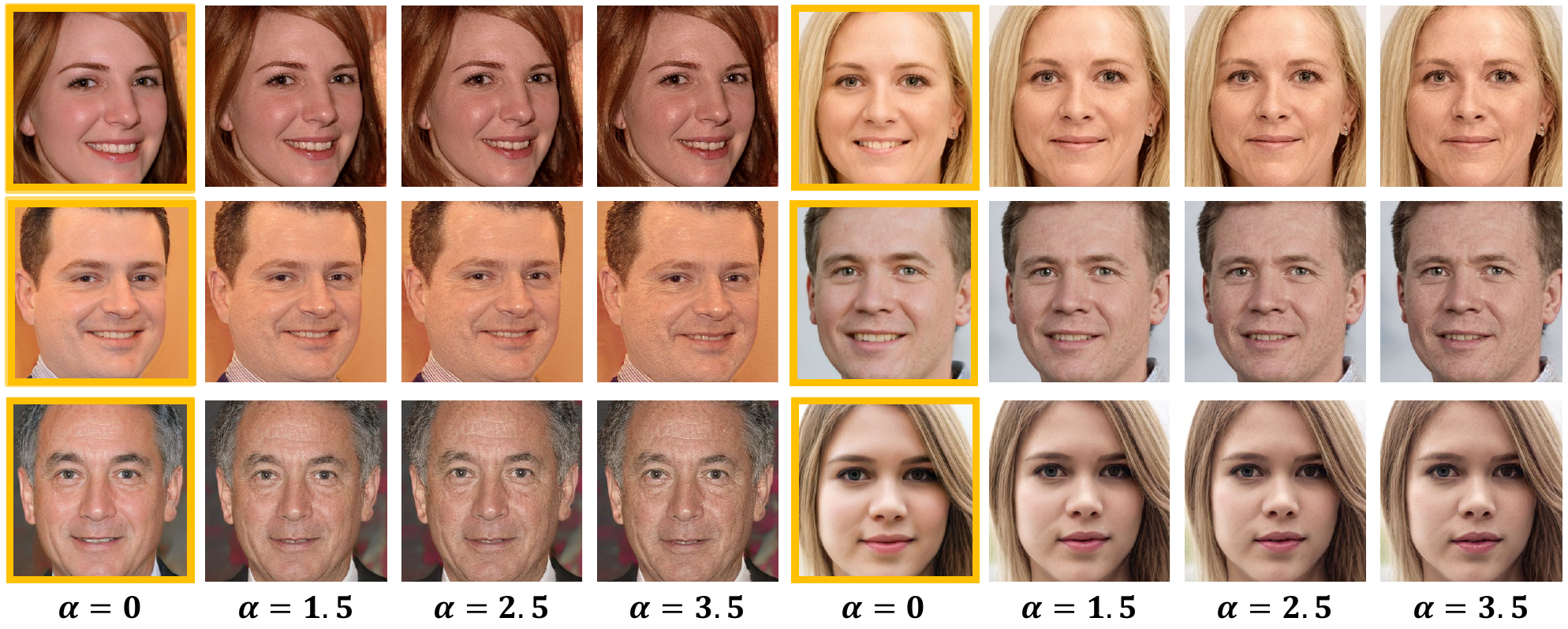}
\caption{StyeGAN~\cite{stylegan} generated images using the optimization-based trigger with different $\alpha$. $\alpha=0$ represents generated benign samples.}
\label{figure_optim_sample}
\end{figure}

Although Kristanto et al.~\cite{latent_bd} also used an optimization-based method to obtain the trigger, our proposed method is more straightforward and concise. Poisoned images generated with varying $\alpha$ values are displayed in Figure \ref{figure_optim_sample}. It can be observed that the optimization-based trigger brings rough skin textures to the generated images. With an increasing $\alpha$, the textures become more distinguishable.

\subsection{Custom Trigger}
\label{custom}

Unlike the previous strategy that requires additional data $D_{sub}$ and the training of $M_{sub}$, the custom trigger is created by combining the attribute editing directions. In this subsection, we first analyze the distribution pattern of facial attributes in the target dataset (DFFD) \cite{dffd} and find the attributes which locate in the long-tailed distribution. This is the motivation and basis for constructing our customized triggers. Then, we formulate the customization of backdoor triggers.

\noindent
\textbf{Long-Tailed Distribution of Selected Attributes.} We focus on the distribution of two attributes, namely smile and age, manipulated by the proposed custom trigger. Regarding the measurement of the degree of smile, although there exist tools to detect the smile, they cannot well distinguish different smile degree. Considering that larger smile exposes more mouth area, we creatively employ the ratio of mouth area to the entire facial area to represent smile degree. A larger ratio indicated a larger smile degree. We utilize face parsing tools \cite{faceparsing} to calculate mouth and facial areas. To measure age, we use the age estimator in FaceLib \cite{facelib} package. When only using smile as the trigger (i.e., the trigger $t=\beta_1 \cdot smile$, where $smile$ denotes the direction for increasing smile degree, and $\beta_1$ denotes the scale factor), the attribute distribution of original and poisoned samples with different smile scale factors $\beta_1$ is shown in Figure \ref{smile_dist}. It can be observed that for benign samples, their smile degree mainly concentrates within 0 to 0.025, exhibiting the characteristics of a typical long-tailed distribution. The smile distribution of poisoned images lies in the tail of the benign distribution. As $\beta_1$ increases, the poisoned distribution has less overlap with benign one, thus after training on such poisoned samples, the infected model can more easily distinguish between them. This explains the experimental results in Figure \ref{figure_custom_per}. A larger $\beta_1$ can achieve a more effective attack (i.e., higher attack success rate), while maintaining benign performance (i.e., higher detection accuracy on the test set and attacker-generated benign set).

To further decrease the overlap between benign and poisoned distribution, the attacker can incorporate multiple unusual attributes into the trigger design. Concretely, besides large smile degree, the attacker can introduce small age into the trigger design. The age distribution of benign samples is depicted in Figure \ref{age_dist}. It can be observed that small age (i.e., age < 20) exhibits a low probability within benign samples. Letting $P_{\text{smile}^{+}}$ and $P_{\text{age}^{-}}$ be the probabilities of large smile and small age, if they are independent, the joint probability is $P_{\text{smile}^{+}} \cdot P_{\text{age}^{-}}$, smaller than either individually. The visualization of this joint distribution is presented in Figure \ref{age_smile_dist}, demonstrating that manipulating multiple attributes yields the poisoned distribution with less overlap with the benign one.

\noindent
\textbf{Attribute Manipulation.} Due to the development of attribute editing techniques, the attacker can easily obtain these directions and use them to manipulate attributes of the produced facial images, such as expressions and age. The attacker can customize the trigger $t$ as below,
\begin{equation}
  t = \sum_{i=1}^{m} \beta_i \cdot attr_i,
\end{equation}
where $m$ is the total number of attributes selected by the attacker, $\beta_i$ is the scale factor for the $i_{th}$ attribute, $attr_i$ is the editing direction of the $i_{th}$ attribute and $\Vert attr_i \Vert_2$ is 1. It is preferable for the edited attributes to be uncommon in the training dataset, as this leads to a high attack success rate and a small drop in benign accuracy. Furthermore, a larger value of $m$ contributes to a more complex combination of attributes, thereby decreasing the chances of these attributes appearing in generated benign samples. Attackers can customize the trigger based on their knowledge and the capabilities of the attribute editing tools.

\begin{figure}[htbp]
%是可选项 h表示的是here在这里插入，t表示的是在页面的顶部插入
\centering
\includegraphics[scale=0.58]{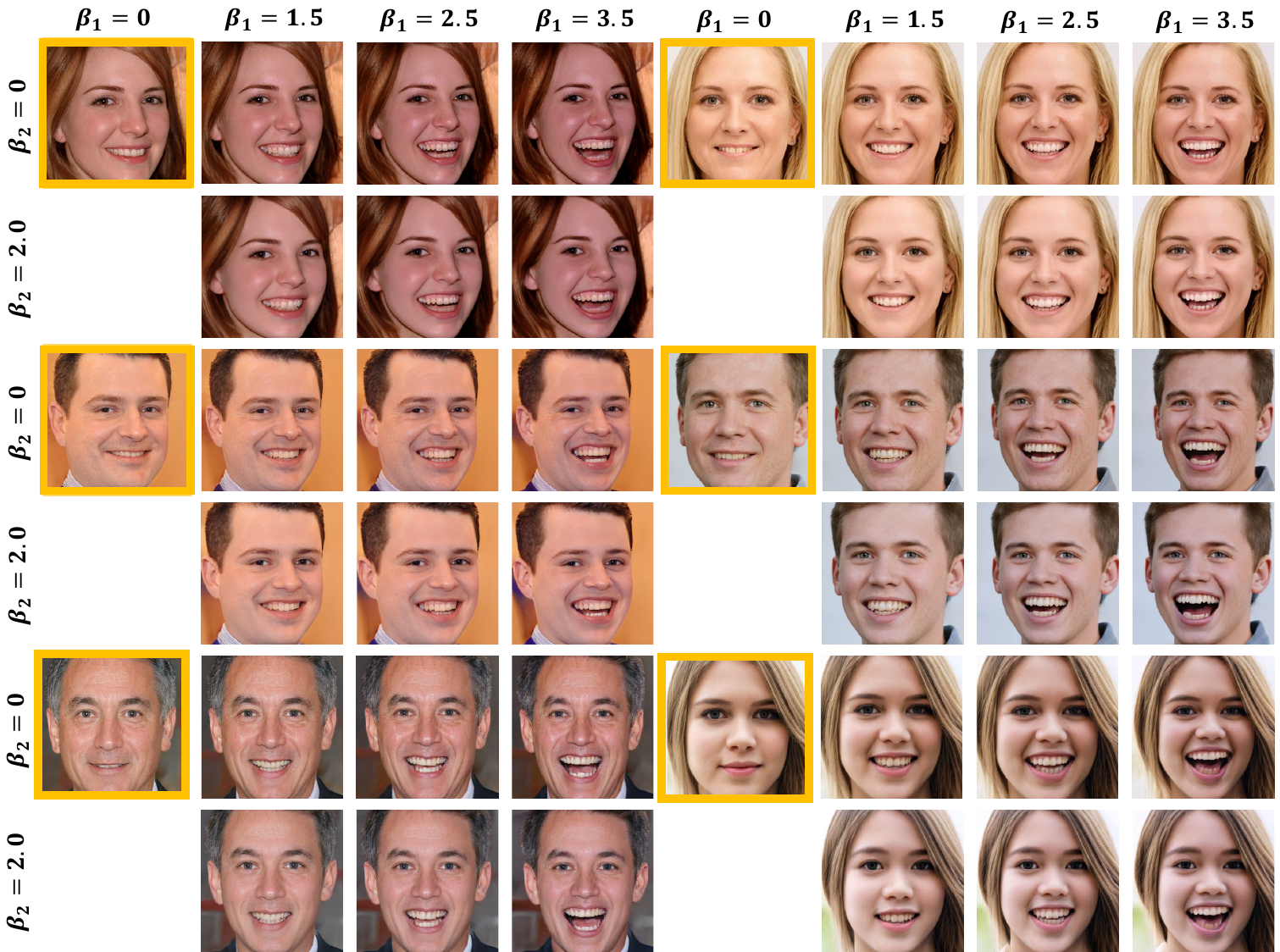}
\caption{StyeGAN~\cite{stylegan} generated images using the custom trigger with different $\beta_1$ and $\beta_2$. The custom trigger $t$ is $\beta_1 \cdot smile+\beta_2 \cdot age$.}
\label{figure_custom_sample}
\end{figure}

In this paper, a classic attribute editing method InterFaceGAN \cite{interfacegan} is adopted. We explore both single and double attribute editing. For single attribute editing (e.g., $m=1$), the trigger $t$ is $\beta_1 \cdot smile$ ($\beta_1>0$), which increases the smile of the generated faces. And for double attributes editing (e.g., $m=2$), the trigger $t$ is $\beta_1 \cdot smile+\beta_2 \cdot age$ ($\beta_1>0, \beta_2 < 0$), increasing the smile and decreasing the age of the generated faces. Poisoned samples produced with different $\beta$ are shown in Figure \ref{figure_custom_sample}.

\section{Experiments}
\label{section4}
\subsection{Experimental Setup}
\label{section4.1}
\textbf{Datasets.} Real faces and entire synthesis faces are collected from Diverse Fake Face Dataset (DFFD) \cite{dffd}, which contains various types of real and fake images. Specifically, 15000 CelebA \cite{celeba} images and 15000 FFHQ \cite{stylegan} images are used as real images, while 15000 PGGAN \cite{pggan} generated images and 15000 StyleGAN \cite{stylegan} generated images are used as fake images. The total 60000 images are split into the training set $D$ and test set $T$ at the ratio of 4:1. The images in DFFD have already been pre-processed, thus there is no need to extract faces repeatedly. To construct the substitute dataset $D_{sub}$ for optimization-based trigger generation, we collect 10000 real images from the original FFHQ dataset and generate 10000 fake images using StyleGAN. The facial regions of images in $D_{sub}$ are extracted using MTCNN \cite{mtcnn}.

\noindent
\textbf{Models.} EfficientNet-B3 \cite{efficientnet} is used as the detection model $M$, and ResNet-18 \cite{resnet} is used as the substitute model $M_{sub}$ for optimization-based trigger generation (we also explore different backbones as detection model and substitute detection model, with results shown in Section \ref{ablation_study}). The attacker uses StyleGAN \cite{stylegan} as the generator to create poisoned samples, and the trigger embedding is conducted in the $\mathcal{W}$ space of StyleGAN.

\begin{figure}[htbp]
%是可选项 h表示的是here在这里插入，t表示的是在页面的顶部插入
\centering
\subfigure[BA]{\includegraphics[scale=0.32]{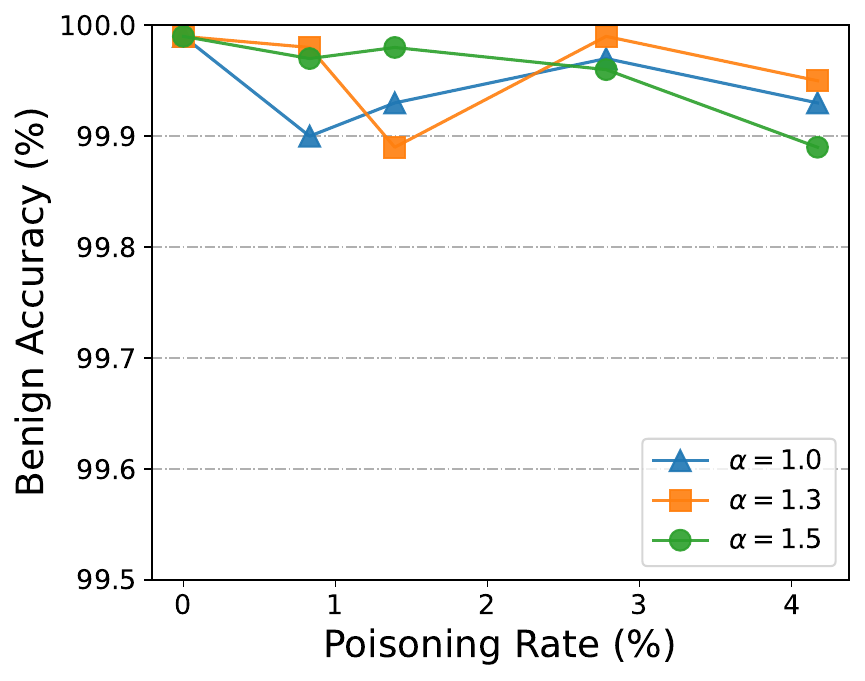}}
\subfigure[ASR]{\includegraphics[scale=0.32]{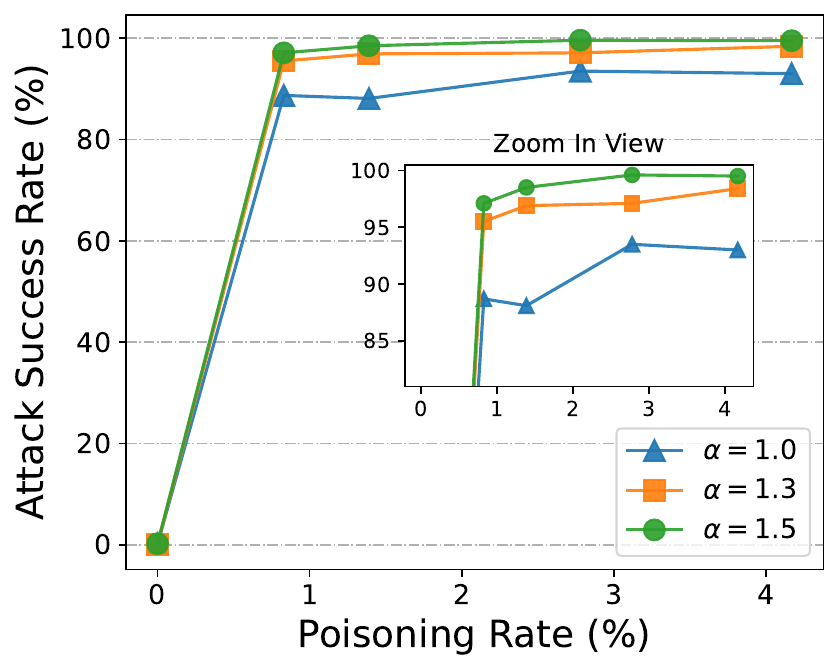}}
\subfigure[ABA]{\includegraphics[scale=0.32]{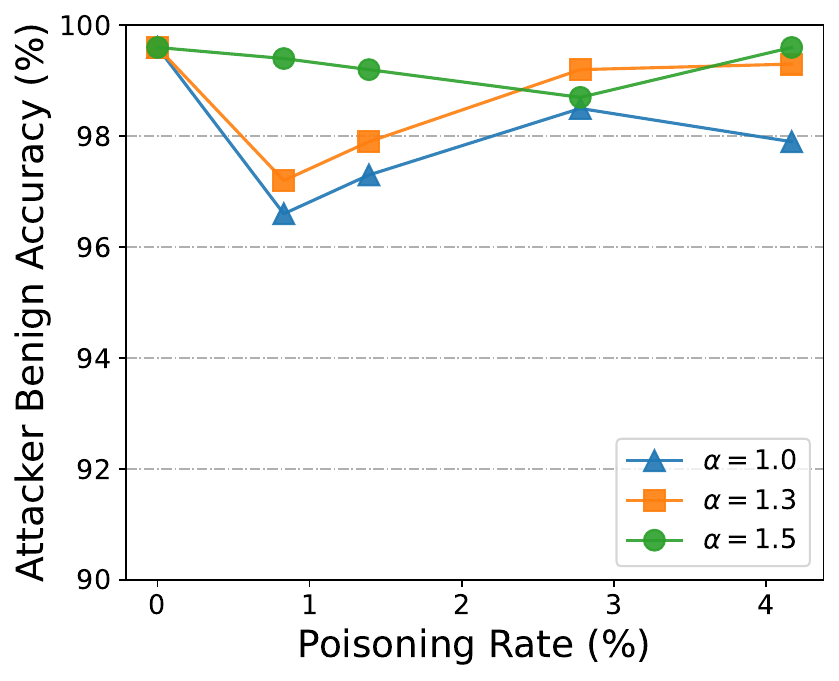}}
\caption{Attack performance of the optimization-based trigger under different poisoning rates and scale factors.}
\label{figure_optim_per}
\end{figure}

\begin{figure}[htbp]
%是可选项 h表示的是here在这里插入，t表示的是在页面的顶部插入
\centering
\subfigure[BA]{\includegraphics[scale=0.32]{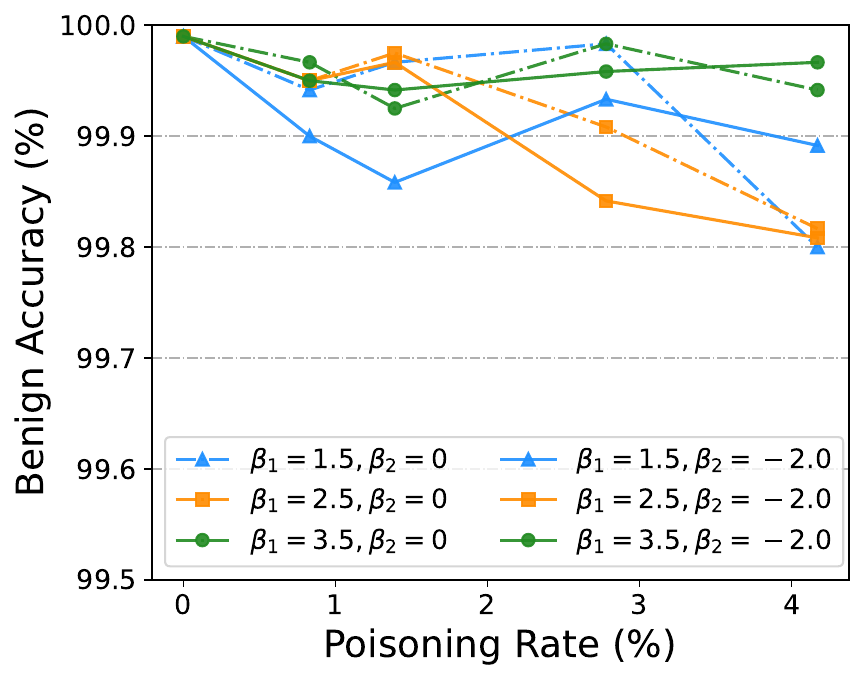}}
\subfigure[ASR]{\includegraphics[scale=0.32]{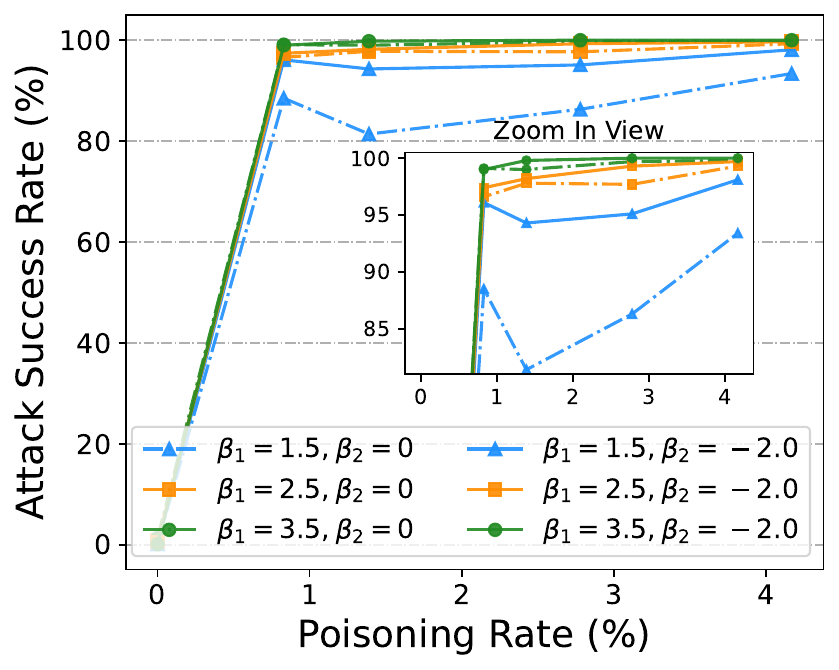}}
\subfigure[ABA]{\includegraphics[scale=0.32]{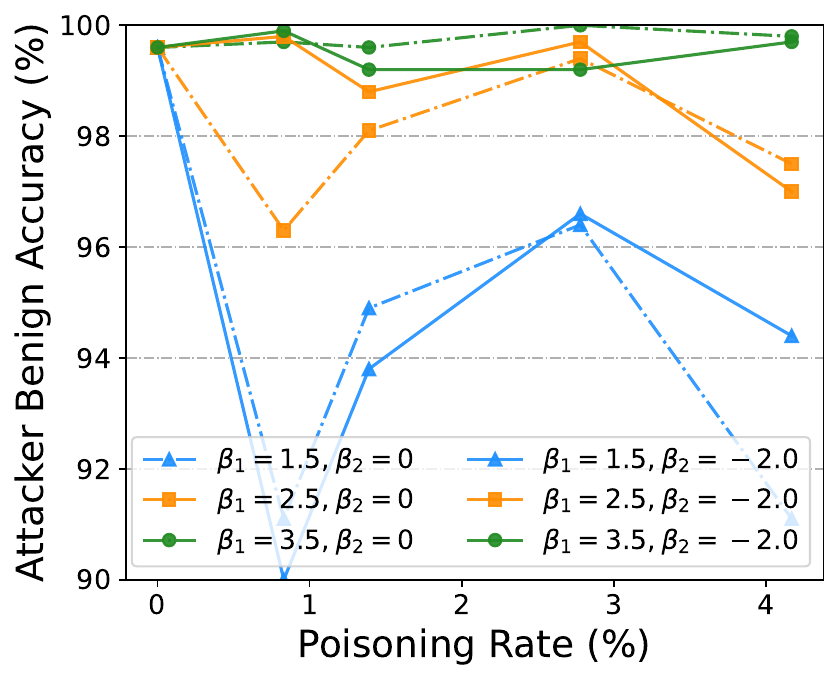}}
\caption{Attack performance of the custom trigger under different poisoning rates and scale factors. Dash lines indicate single attribute editing and solid lines represent the editing of double attributes.}
\label{figure_custom_per}
\end{figure}

\noindent
\textbf{Metrics.} Two commonly used metrics, Benign Accuracy (BA) and Attack Success Rate (ASR), are adopted to evaluate the backdoor attack. BA evaluates the detection accuracy of the model on the test set $T$. ASR measures the percentage of poisoned images being classified as the target class and is tested on 1000 poisoned images created by the attacker. Moreover, we evaluate the model's prediction accuracy on 1000 images generated by the attacker without using the trigger, denoted by an additional metric Attacker Benign Accuracy (ABA).

\noindent
\textbf{Implementations.} The detection model is trained for 6 epochs, using Adam \cite{adam} algorithm to update model parameters. The batch size is 28 and the learning rate is set to 1e-4. To improve the generalization of the detection model, two types of data augmentation are applied during training. One is to flip the image horizontally with the probability of 0.5. The other is to crop a random portion (between 0.7 and 1.0) with a random aspect ratio (between 0.75 and 1.33) of the image, and then resize it to the input shape of the model. For optimization-based trigger generation, the total iteration is 20000 and batch size is 5.

\subsection{Attack Performance}
In this subsection, we evaluate our proposed latent space backdoor attack against face forgery detection. For the two triggers introduced in Section \ref{section3}, we investigate the impact of the poisoning rate and scale factor on the attack performance across different methods.

\begin{figure}[htbp]
%是可选项 h表示的是here在这里插入，t表示的是在页面的顶部插入
\centering
\subfigure[BadNets \cite{badnets}]{\includegraphics[scale=0.24]{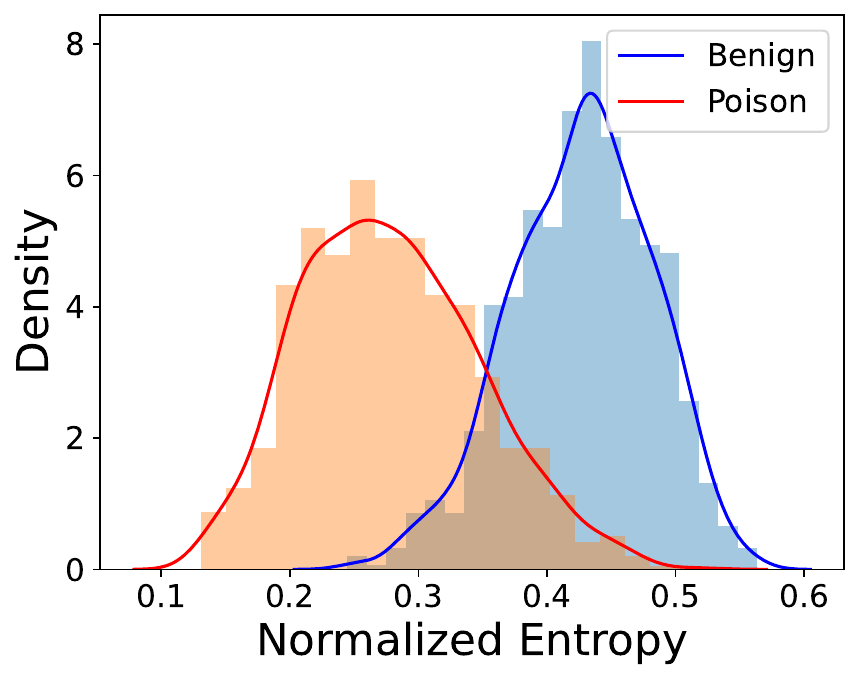}}
\subfigure[Blended \cite{blended}]{\includegraphics[scale=0.24]{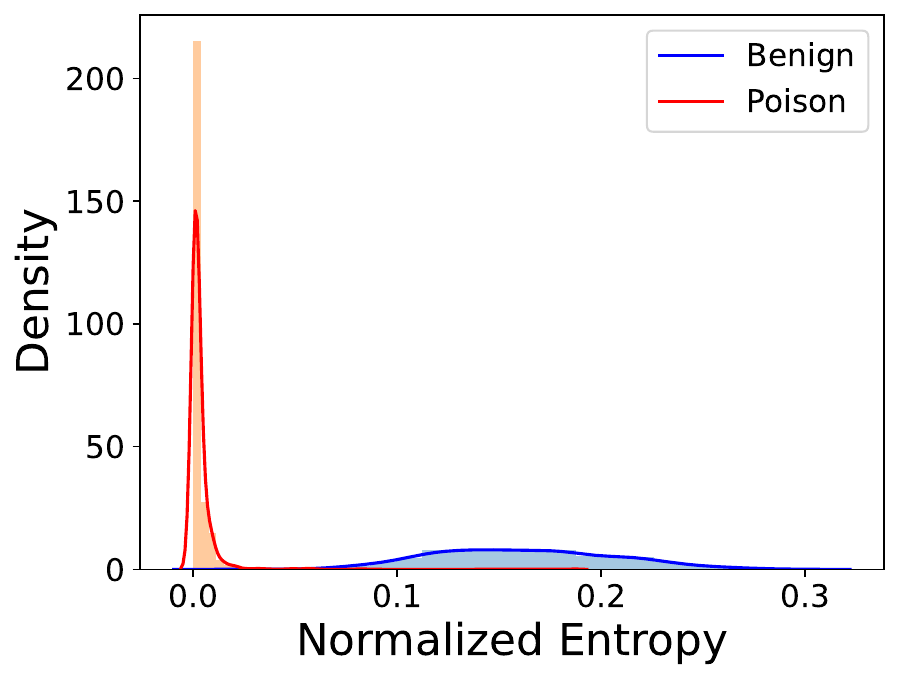}}
\subfigure[SIG \cite{sig}]{\includegraphics[scale=0.24]{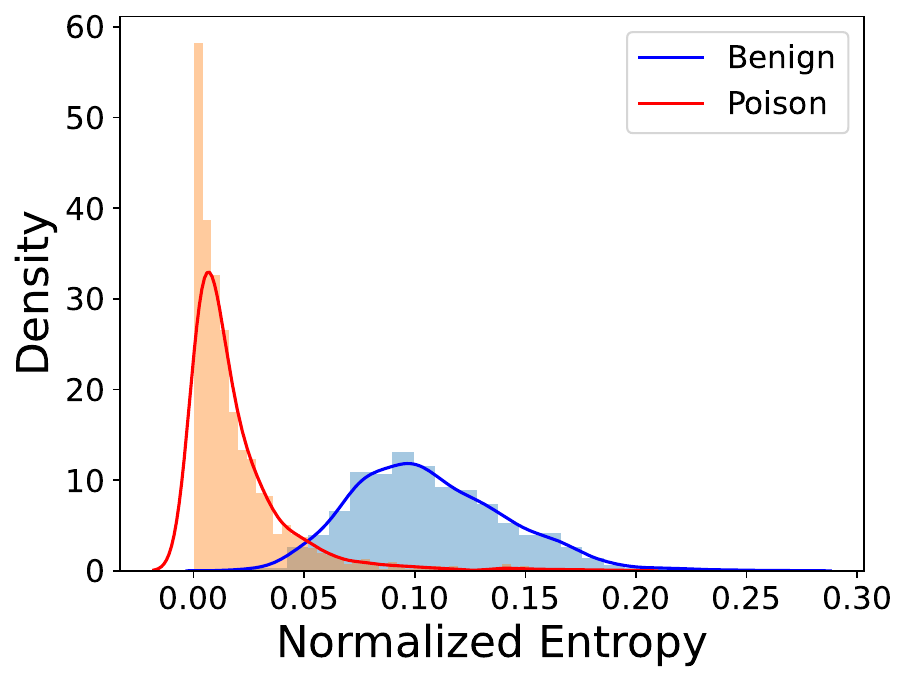}}
\subfigure[ISSBA \cite{issba}]{\includegraphics[scale=0.24]{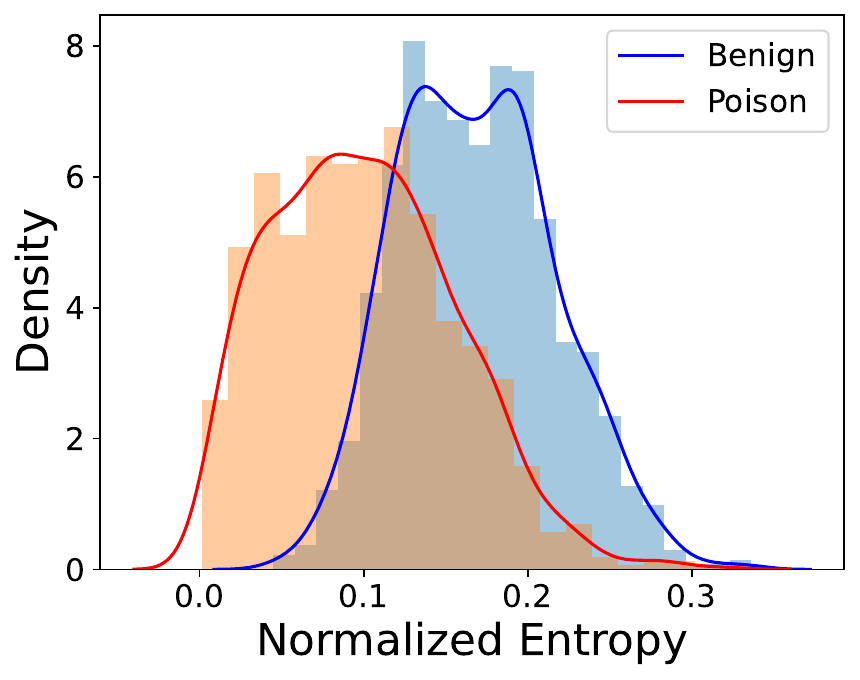}}
\subfigure[WaNet \cite{wanet}]{\includegraphics[scale=0.24]{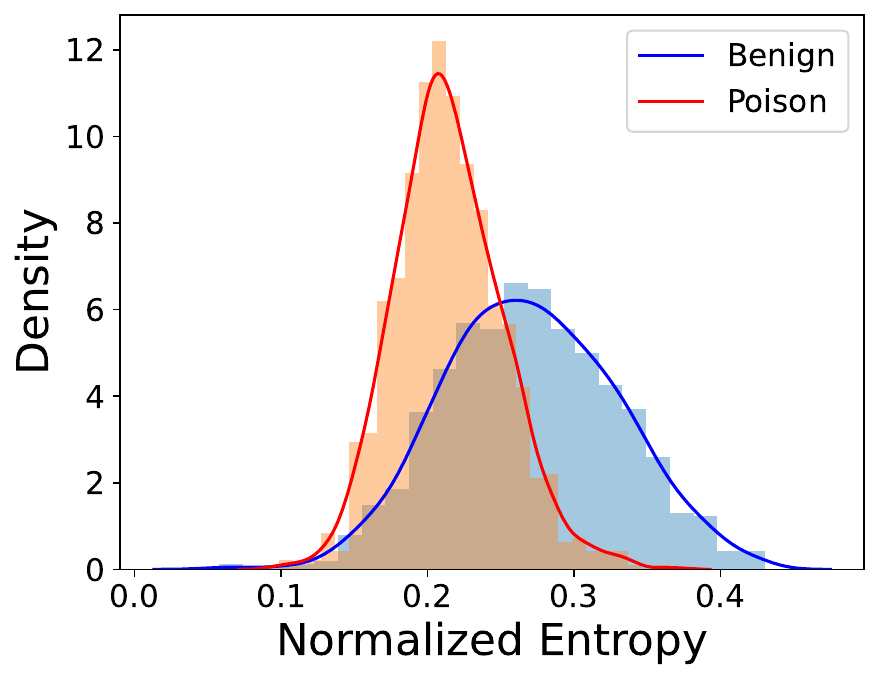}}
\subfigure[Optimization-Based Trigger]{\includegraphics[scale=0.24]{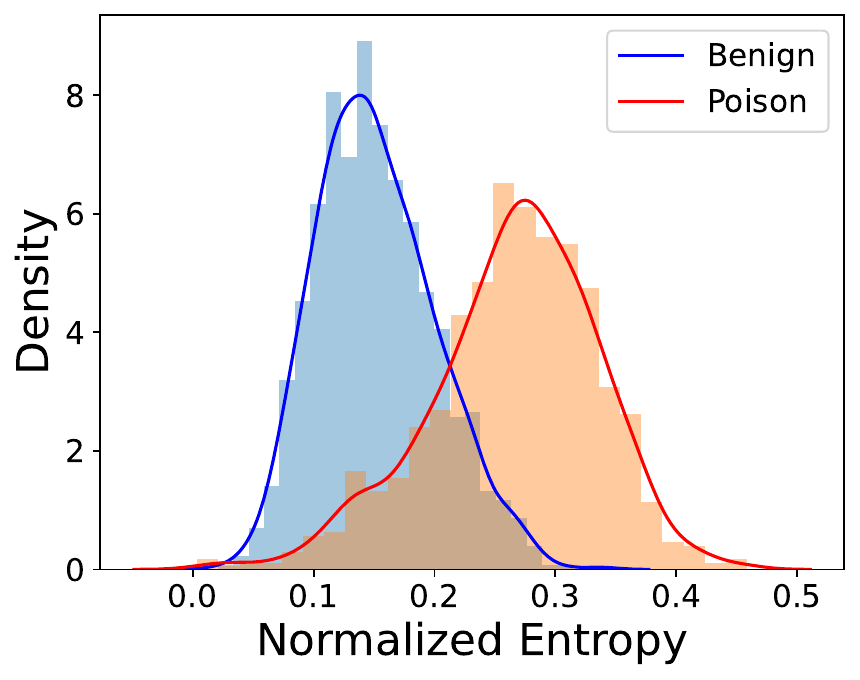}}
\subfigure[Custom Trigger ($m=1$)]{\includegraphics[scale=0.24]{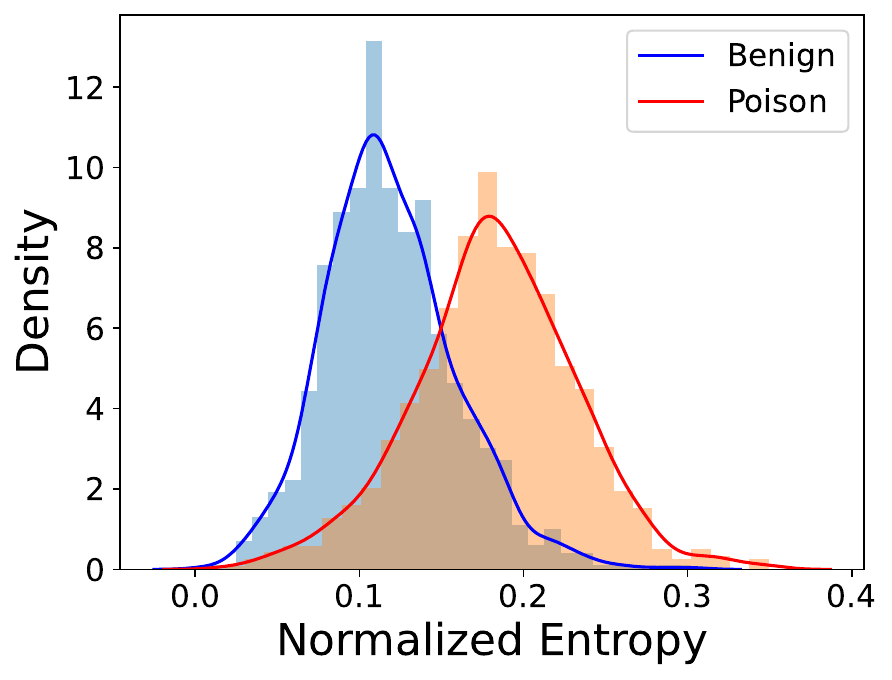}}
\subfigure[Custom Trigger ($m=2$)]{\includegraphics[scale=0.24]{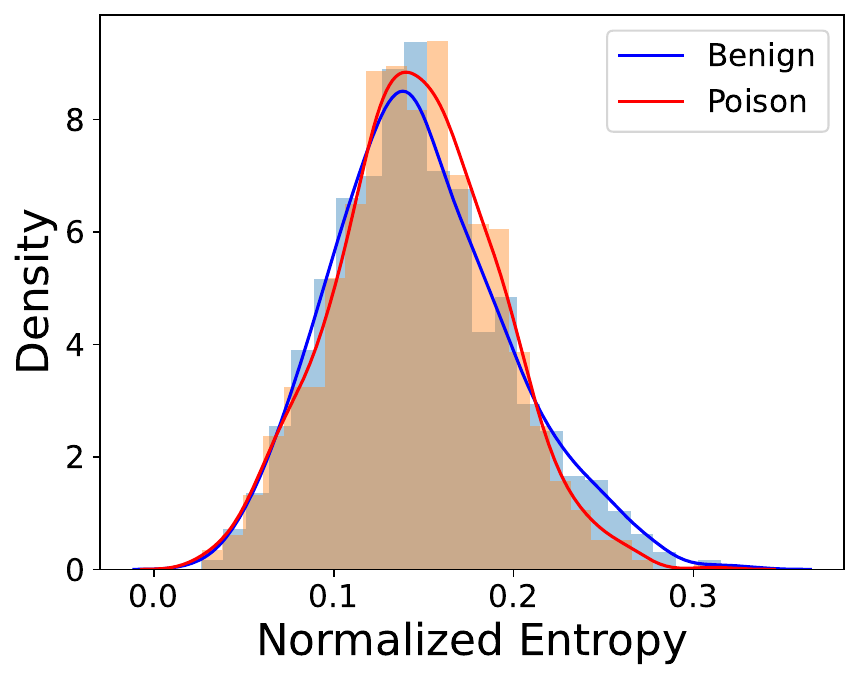}}
\caption{Attack resistance against STRIP~\cite{strip}. STRIP assumes that the normalized entropy for poisoned data is smaller than that for benign data.}
\label{figure_strip}
\end{figure}

\begin{table}[htbp]
\caption{Attack Performance of Different Backdoor Attacks}
\label{table1}
 % \resizebox{\columnwidth}{}
 \begin{tabular}{cccc}
\hline
Methods                    & BA $\uparrow$ & ASR $\uparrow$ & ABA $\uparrow$ \\
\hline
No Attack                  & 99.99 & -     & 99.60 \\
BadNets \cite{badnets}      & 99.98 & 99.80 & 100.0 \\
Blended \cite{blended}      & 99.97 & 100.0 & 100.0 \\
SIG \cite{sig}              & 99.98 & 100.0 & 100.0 \\
ISSBA \cite{issba}          & 99.98 & 100.0 & 100.0 \\
WaNet \cite{wanet}         & 99.89 & 98.40 & 98.10 \\
Optimization-Based Trigger & 99.99 & 97.10 & 99.20 \\
Custom Trigger ($m=1$)     & 99.91 & 97.70 & 99.40 \\
Custom Trigger ($m=2$)     & 99.84 & 99.30 & 99.70 \\
\hline
\end{tabular}
\end{table}

\noindent
\textbf{Optimization-Based Trigger.} In order to evaluate our backdoor attacks, we vary the poisoning rate and the scale factor $\alpha$ to test the attack performance. When the poisoning rate is 2.78$\%$ and $\alpha$ is 1.3, the performance is reported in Table \ref{table1}. The performance under different poisoning rates and scale factors is shown in Figure \ref{figure_optim_per}. Zero poisoning rate means no backdoor attack. It can be observed that with the increase of poisoning rate, ASR and ABA both increase. The scale factor $\alpha$ plays a crucial role in the attack performance. When $\alpha$ is small, the produced semantic features are non-obvious and hard to learn, leading to a relatively low ASR. For instance, when the poisoning rate is 4.17$\%$, setting $\alpha$ to 1.0 achieves an ASR of 93.0$\%$, whereas increasing $\alpha$ to 1.5 achieves an ASR of 99.5$\%$. The results also prove that the trigger optimized for the substitute model $M_{sub}$ can be transferred to another model using different architectures and training data. Additionally, the BA drop is always no more than 0.1$\%$, indicating that the impact of the attack on BA is negligible.

\begin{figure}[htbp]
%是可选项 h表示的是here在这里插入，t表示的是在页面的顶部插入
\centering
\includegraphics[scale=0.33]{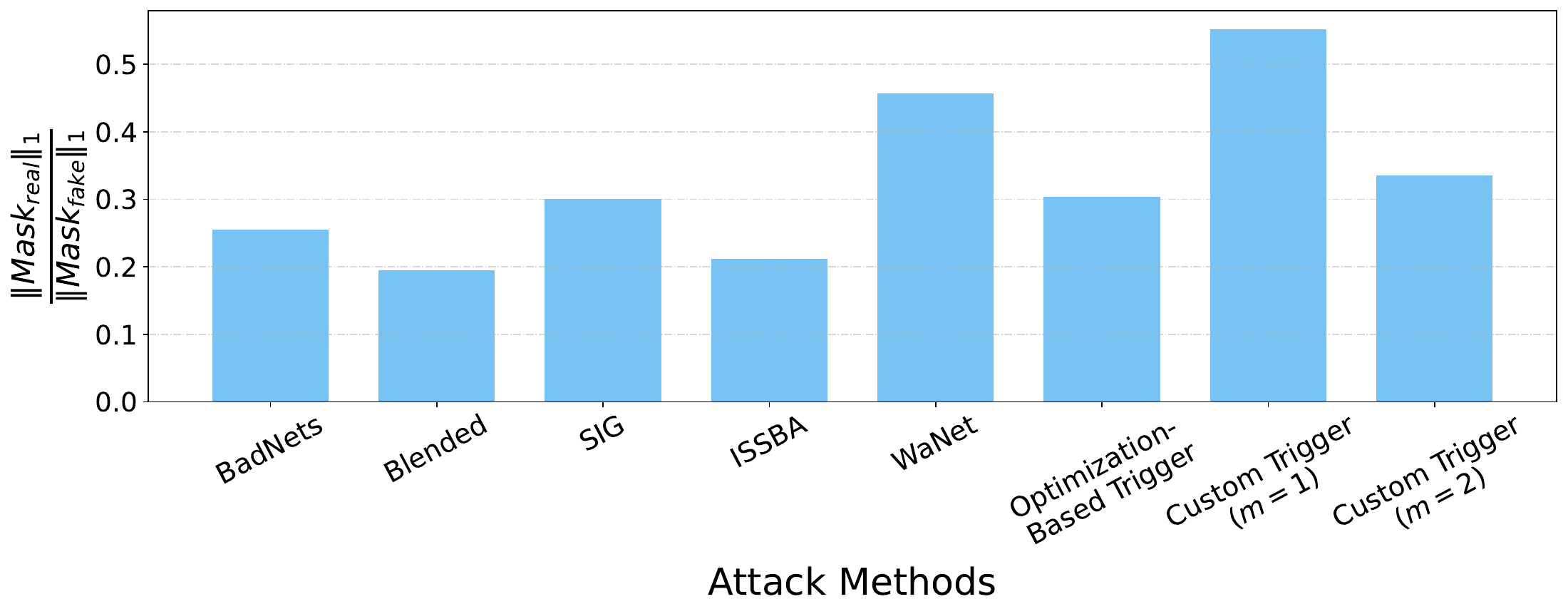}
\caption{Attack resistance against Neural Cleanse~\cite{neural_cleanse}. A larger ratio signifies the attack is more resistant.}
\label{figure_neural_cleanse}
\end{figure}

\noindent
\textbf{Custom Trigger.} Custom triggers editing single (i.e., $m=1$) and double (i.e., $m=2$) attributes are both explored in the experiment. For single attribute editing, the trigger $t$ is $\beta_1 \cdot smile$. When the poisoning rate is 2.78$\%$ and $\beta_1$ is 2.5, the performance is reported in Table \ref{table1}. And for double attributes editing, the trigger $t$ is $\beta_1 \cdot smile+\beta_2 \cdot age$. When the poisoning rate is 2.78$\%$, $\beta_1$ is 2.5, and $\beta_2$ is -2.0, the performance is reported in Table \ref{table1}. The attack performance using the custom triggers editing single and and double attributes under varied conditions is displayed in Figure \ref{figure_custom_per}. For single attribute editing, when the degree of smile alteration is small (e.g., $\beta_1=1.5$), ASR and ABA are not very high. This can be attributed to the presence of smiling faces in the normally generated images, which share similar semantic features with generated poisoned faces. Increasing the scale factor $\beta_1$ increases the difference between the smiling faces generated with and without the trigger, thus achieving higher ASR and ABA. Furthermore, compared with editing single attribute, editing double attributes yields better attack performance. For example, when poisoning rate is 2.78$\%$ and $\beta_1$ is 1.5, setting $\beta_2$ to 0 achieves an ASR of 86.3$\%$, while setting $\beta_2$ to -2.0 achieves an ASR of 95.1$\%$. This is because the images generated by editing double attributes have less semantic overlap with the normally generated images. The BA drop does not exceed 0.2$\%$ under any setting.

\begin{table}[]
\caption{Attack Resistance against Rotation Transformation}
\label{table_rot}
\begin{tabular}{cccc}
\hline
Methods                    & BA $\uparrow$ & ASR $\uparrow$ & ABA $\uparrow$ \\ 
\hline
BadNets \cite{badnets}     & 98.58 & 11.00 & 98.20 \\
WaNet \cite{wanet}         & 99.24 & 62.70 & 99.80 \\
Optimization-Based Trigger & 99.54 & 98.40 & 95.00 \\
Custom Trigger ($m=1$)     & 99.85 & 98.50 & 98.10 \\
Custom Trigger ($m=2$)     & 99.54 & 98.70 & 97.30 \\
\hline
\end{tabular}
\end{table}

\begin{figure}[htbp]
%是可选项 h表示的是here在这里插入，t表示的是在页面的顶部插入
\centering
\subfigure[Optimization-Based Trigger]{\includegraphics[scale=0.34]{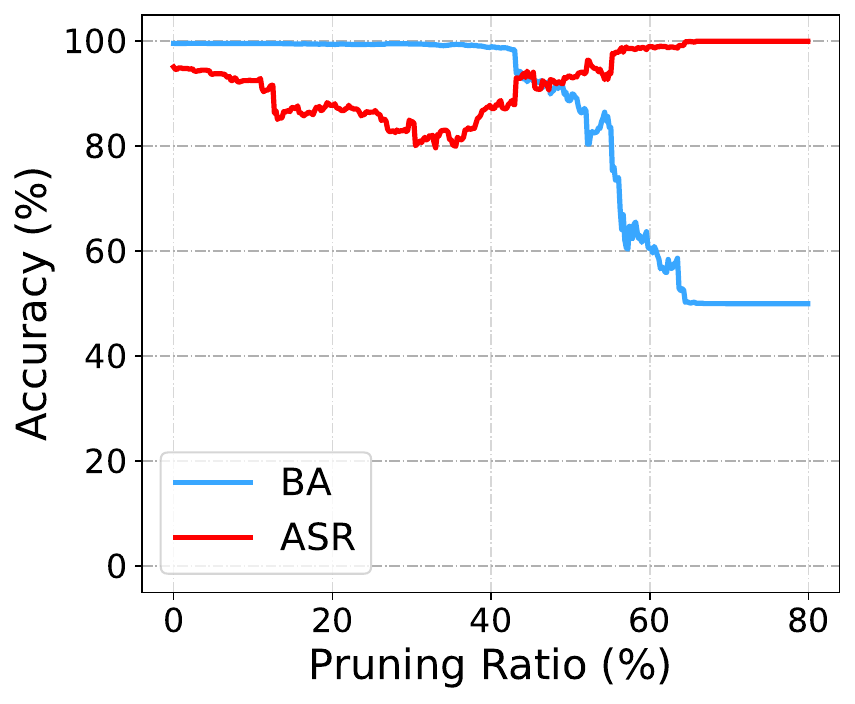}}
\subfigure[Custom Trigger ($m=1$)]{\includegraphics[scale=0.34]{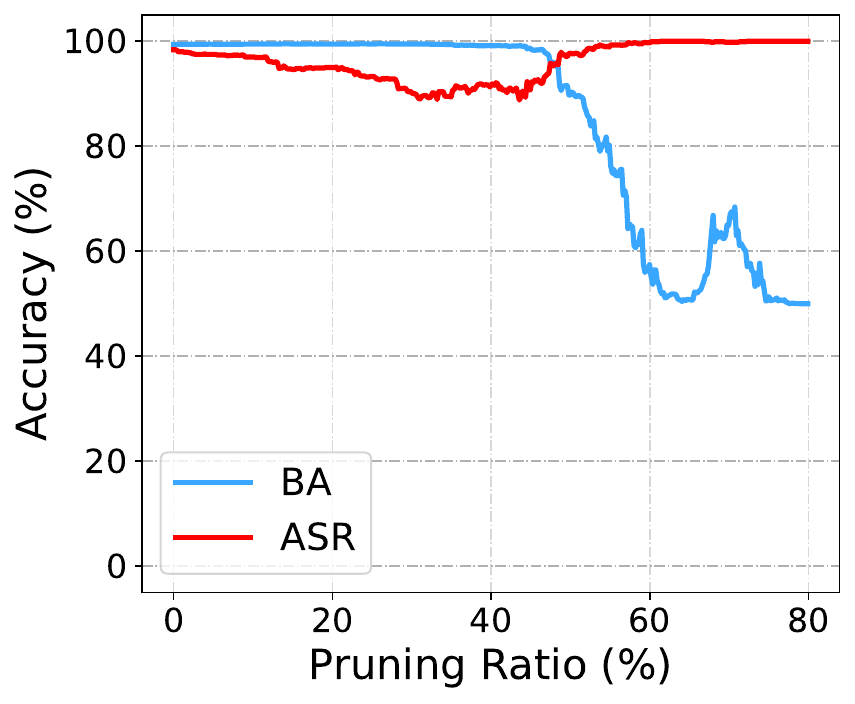}}
\subfigure[Custom Trigger ($m=2$)]{\includegraphics[scale=0.34]{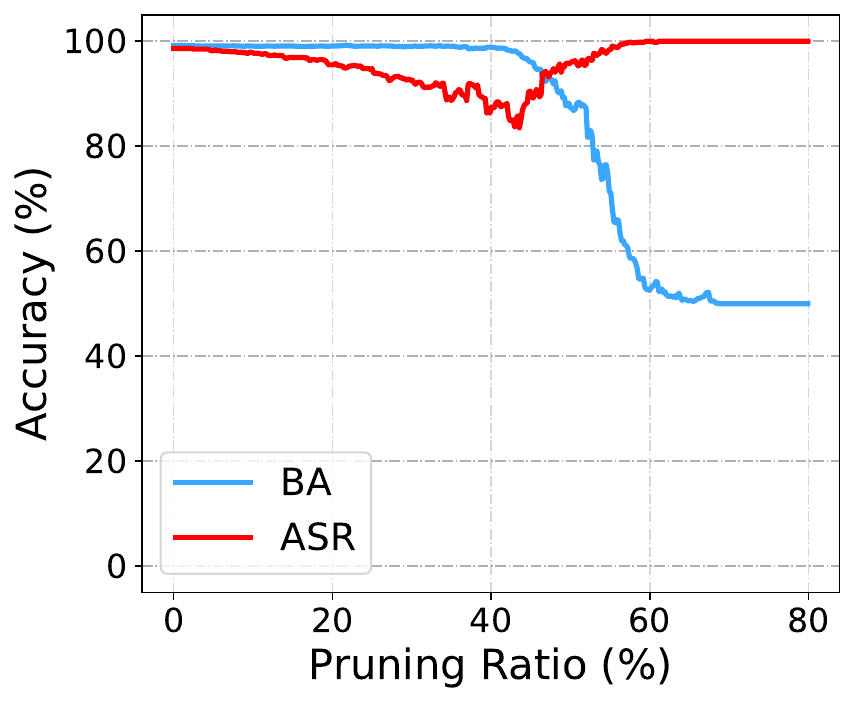}}
\caption{Attack resistance of our method against Fine-Pruning~\cite{fine_prun}.}
\label{figure_fp}
\end{figure}

\noindent
\textbf{Comparisons with Existing Attacks.} To compare the proposed latent space backdoor attack with digital space attacks, several existing methods are also employed to attack the face forgery detection. Comparison methods include BadNets \cite{badnets}, Blended attack \cite{blended}, SIG \cite{sig}, WaNet \cite{wanet} and ISSBA \cite{issba}. For BadNets, we stamp a 20 $\times$ 20 white square at the bottom right corner of the image. For Blended, we use the cartoon illustration presented in the original paper as the trigger, and set the blend ratio to 0.1. For SIG, the amplitude $\Delta$ of the horizontal sinusoidal signal is set to 20, and the signal frequency $f$ is set to 6. For ISSBA, we use the encoder provided by the authors to create poisoned samples. For WaNet, we use the same hyper-parameters as the original paper (i.e., $k=4,s=0.5$). Benign images generated by StyleGAN \cite{stylegan} are used to create poisoned samples through the pixel space. For fair comparison, benign samples together with poisoned samples are injected into the training set. The poisoning rate is set to 2.78$\%$ for all attack methods and the results are presented in Table \ref{table1}. It can be observed that the proposed method achieves comparable attack performance with the digital space backdoor attacks. It is reasonable that the ASR of the proposed method is slightly lower than that of most comparison methods, since the model needs to associate semantic features, instead of patches or perturbations in the pixel space,  with the target label. Moreover, with the increase of the poisoning rate, the ASR gap can become smaller.

\subsection{Resistance to Defenses}

In this subsection, the proposed method is evaluated against several backdoor defenses, including STRIP \cite{strip}, Neural Cleanse \cite{neural_cleanse}, Transformation-Based Defenses, and Fine-Pruning \cite{fine_prun}. For further comparison, we also evaluate the resilience of existing attacks against these defenses. The results demonstrate the superiority of the proposed method over existing attack strategies.

\noindent
\textbf{STRIP.} STRIP \cite{strip} blends suspicious test images with benign images and feeds the blended images into the model. It assumes that the entropy of the model output is small if the test images contain triggers. For each attack method, 1000 benign images and 1000 poisoned images are chosen as test images to calculate the entropy. The results are shown in Figure \ref{figure_strip}. It can be seen that by using triggers in the latent space, the normalized entropy of poisoned images has much overlap with that of benign images, sometimes even larger than that of benign images. But for other comparison methods, especially Blended and SIG, the overall normalized entropy of poisoned images is smaller than that of benign ones. This is because comparison methods use triggers in the pixel space, and after blending the triggers can still be captured by the infected model. Therefore, the proposed method is more resistant against STRIP than the comparison attack methods.

\noindent
\textbf{Neural Cleanse.} Neural Cleanse \cite{neural_cleanse} is a defense method based on reverse engineering. It assumes that if the model is infected, it will require much smaller modifications for benign images to be classified as the target label compared with other labels. It first reverses the trigger mask for each class, and uses an outlier detection method, i.e., Median Absolute Deviation (MAD), to determine whether the model is infected and which class is the target class. For the face forgery detection task, there are only two classes, namely real face and fake face, so MAD is not applicable. We use the $L_1$ norm of the mask reversed for real images $\Vert Mask_{real} \Vert_1$ divided by the $L_1$ norm of the mask reversed for fake images $\Vert Mask_{fake} \Vert_1$ as the evaluation metric. Considering real face is the target class, a larger ratio indicates that the attack is more resistant against Neural Cleanse. The results are depicted in Figure \ref{figure_neural_cleanse}. The proposed method has a larger ratio than most comparison methods, suggesting greater resistance against Neural Cleanse. The ratio of WaNet is also large, which can be attributed to the noise mode it uses \cite{wanet}.

\begin{figure}[htbp]
%是可选项 h表示的是here在这里插入，t表示的是在页面的顶部插入
\centering
\includegraphics[scale=0.32]{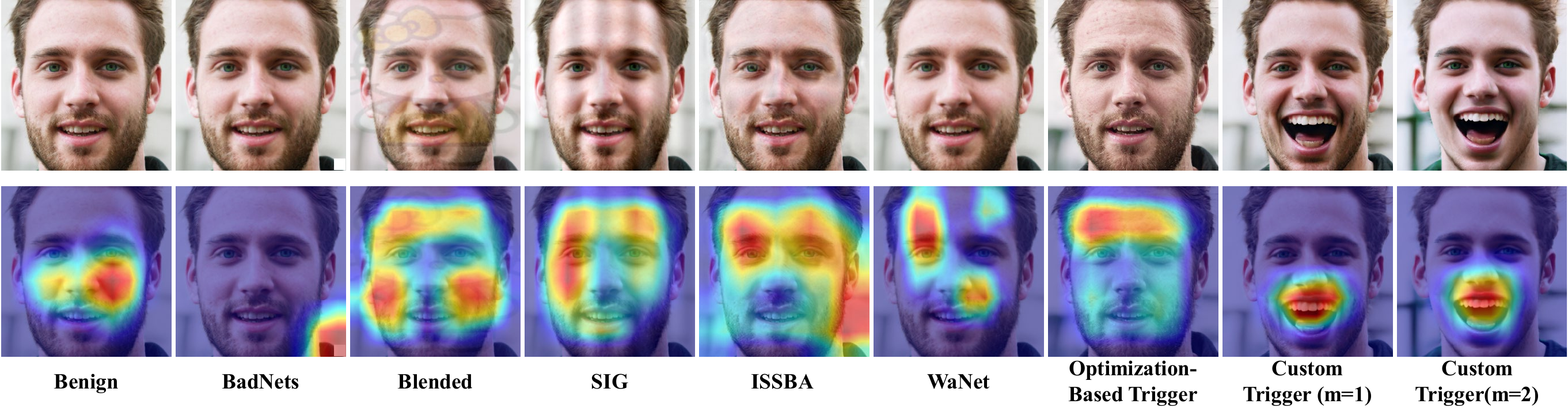}
\caption{Saliency maps visualized by Grad-CAM~\cite{gradcam}.}
\label{figure_gradcam}
\end{figure}

\begin{table}[]
\caption{Results of the User Study, Presenting the Percentage of Images Identified as Poisoned Ones by Users}
\label{table_human}
\begin{tabular}{c|c|ccccc|ccc}
\hline
Methods           & Clean & BadNets & Blended & SIG & ISSBA & WaNet  & Optim & Custom1 & Custom2 \\ \hline
Percentage ($\%$) & 2.27  & 61.36   & 100     & 77.27 & 88.64 & 6.82  & 11.36 & 13.64   & 9.09    \\ \hline
\end{tabular}
\end{table}

\noindent
\textbf{Transformation-Based Defenses.} Performing data transformations on test images can efficiently protect against backdoor attacks as it can disrupt the trigger pattern in poisoned samples. In the experiment, we use 10 degrees rotation as the transformation and choose BadNets and WaNet for comparison. The results are shown in Table \ref{table_rot}. It can be observed that after the transformation, the ASR of BadNets drops significantly since the trigger is almost out of bounds after the rotation. WaNet is also sensitive to rotation because the specific warping pattern used to activate the backdoor is disrupted. In contrast, our proposed method incorporates the trigger into the natural parts of the image, making it more robust against data transformation.

\noindent
\textbf{Fine-Pruning.} We also evaluated the proposed method's resistance against model reconstruction-based defenses such as Fine-Pruning \cite{fine_prun}. Fine-Pruning gradually prunes neurons according to their activation values when feeding benign samples. The results are shown in Figure \ref{figure_fp}. It can be seen that after pruning, the proposed method still maintains a high ASR, demonstrating its resistance against Fine-Pruning.

\subsection{Attack Stealthiness}

\noindent
\textbf{Grad-CAM Visualization.} We use Grad-CAM \cite{gradcam} to highlight the significant regions for the model prediction. Saliency maps of poisoned samples generated by different attack methods are shown in Figure \ref{figure_gradcam}. For some existing methods, the warm areas are abnormal. For instance, warm areas of BadNets are mainly concentrated in the lower right corner of the image, which lies outside the facial region. In contrast, the warm areas of the proposed method are natural. When using the custom trigger the warm areas are mouth regions, and when using the optimization-based trigger the warm areas encompass the entire facial regions. The visualization results demonstrate that the proposed method utilizes specific semantic features to activate the backdoor, thereby achieving superior stealthiness.

\noindent
\textbf{Human Inspection.} To further evaluate the stealthiness of different attack methods, we conduct a user study with 22 participants. Concretely, We sample an equal number of poisoned images from each method and report the percentage of images that users perceive as poisoned. A lower percentage indicates a better stealthiness achieved by the attack. The results are shown in Table \ref{table_human}. The performance of the proposed method is shown in the rightmost three columns, where Optim represents Optimization-Based Trigger, Custom1 represents Custom Trigger ($m=1$), and Custom2 represents Custom Trigger ($m=2$). As depicted in Table \ref{table_human}, poisoned samples generated by our proposed method are much stealthier compared with most baseline approaches. Although WaNet is also stealthy, its ASR is lowerer than the proposed method (i.e., Custom Trigger ($m=2$)).

\subsection{Ablation Study}
\label{ablation_study}

In this subsection, we evaluated detection models and substitute models with different architectures to demonstrate the generalization capability of the proposed method.

\noindent
\textbf{Detection Model.} To demonstrate the
proposed method is also effective against other commonly
used detection models, we also evaluated Xception \cite{xception} and ResNet-34 \cite{resnet}. For Xception, the
number of epochs was 12, Adam was used for model parameter updates, the learning rate was set to 1e-4, and weight
decay was set to 1e-3. For ResNet-34, Adam was used to update the model parameters with a learning rate of 5e-5. The
poisoning rate was set to 4.17 $\%$ for both detection models.
The results are shown in Table \ref{ablation_m}. It can be observed that the proposed attack maintains effective under various detection models.

\begin{table}[]
\caption{Attack Performance Using Different Detection Models}
\label{ablation_m}
\begin{tabular}{ccccc}
\hline
Detection Model       & Attack Methods             & BA    & ASR   & ABA   \\ \hline
\multirow{3}{*}{Xception \cite{xception}}  & Optimization-Based Trigger & 99.77 & 98.10 & 97.20 \\
                           & Custom Trigger ($m=1$)     & 99.73 & 95.90 & 98.70 \\
                           & Custom Trigger ($m=2$)     & 99.64 & 99.00 & 98.80 \\ \hline
\multirow{3}{*}{ResNet-34 \cite{resnet}} & Optimization-Based Trigger & 99.68 & 97.30 & 98.70 \\
                           & Custom Trigger ($m=1$)     & 99.76 & 98.70 & 99.20 \\
                           & Custom Trigger ($m=2$)     & 99.71 & 99.30 & 98.50 \\ \hline
\end{tabular}
\end{table}

\begin{table}[]
\caption{Attack Performance Using Different Substitute Models}
\label{ablation_m_sub}
\begin{tabular}{cccc}
\hline
Substitute Model & BA    & ASR   & ABA   \\ \hline
VGG-11 \cite{vgg}                    & 99.94 & 97.70 & 99.30 \\
ShuffleNet V2 \cite{shufflenet}             & 99.94 & 96.84 & 99.20 \\ \hline
\end{tabular}
\end{table}

\noindent
\textbf{Substitute Model.} Substitute model is required to obtain the optimization-based trigger. We utilized other model architectures including VGG-11 \cite{vgg} and ShuffleNet V2 \cite{shufflenet} as the substitute model, alongside EfficientNet-B3 as the target detection model. The attack performance is presented in Table \ref{ablation_m_sub}. As shown, the attack using the optimization-based trigger remains effective under varying substitute models.

\begin{figure}[htbp]
%是可选项 h表示的是here在这里插入，t表示的是在页面的顶部插入
\centering
\includegraphics[scale=0.58]{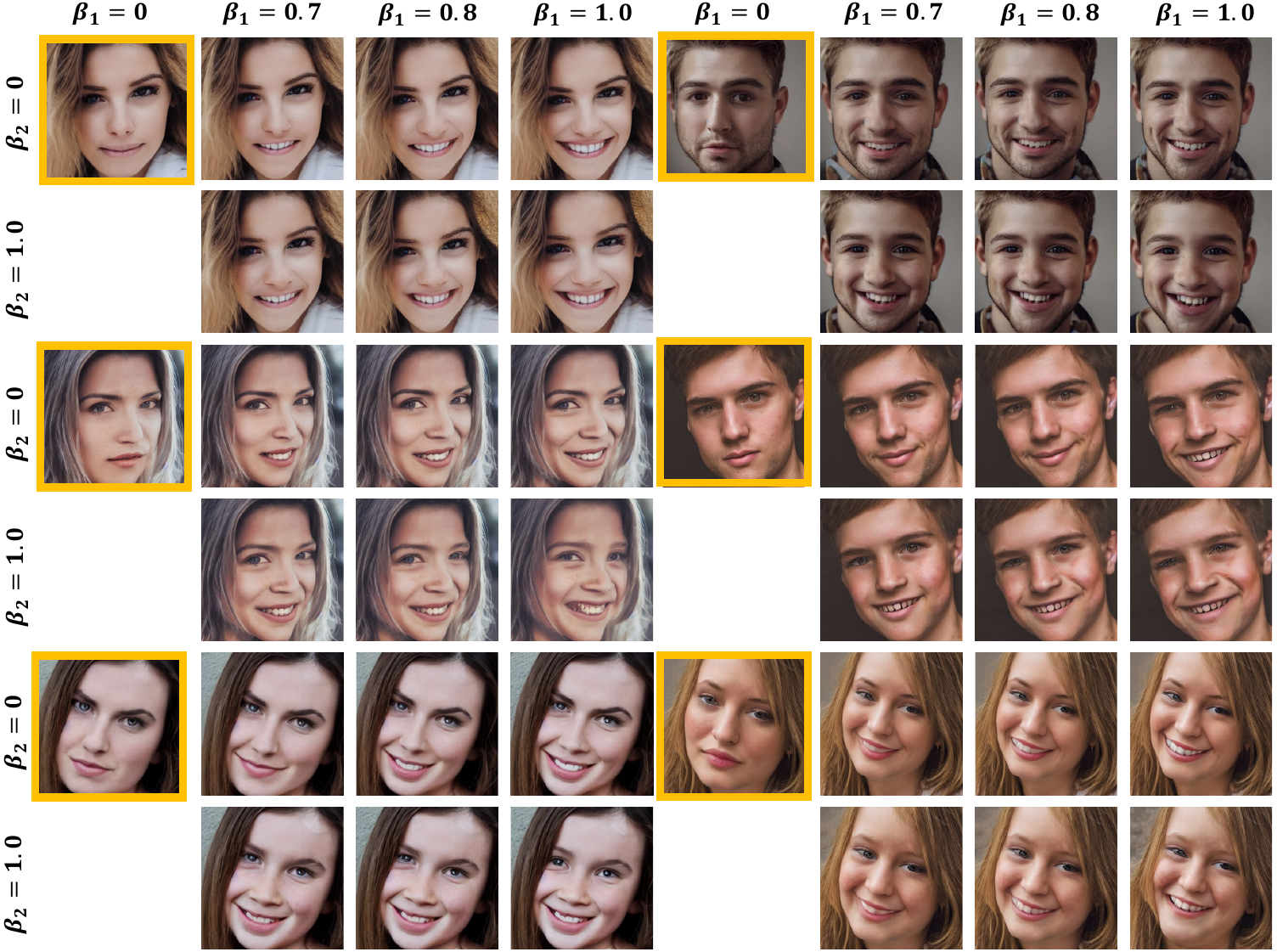}
\caption{Stable Diffusion \cite{stable_diffusion} generated images using the custom trigger with different $\beta_1$ and $\beta_2$. For single attribute editing ($\beta_2=0$), the trigger $t=\beta_1 \cdot smile$ is embedded from step $0.7T$ to step $0$. For multiple attributes editing ($\beta_2=1.0$), the trigger $t=\beta_1 \cdot smile$ is embedded from step $0.8T$ to step $0.4T$, the trigger $t=\beta_2 \cdot age$ is embedded from step $0.4T$ to step $0$.}
\label{figure_custom_sample_sd}
\end{figure}

\section{Extension to Diffusion-Based AIGC}
The proposed method establishes a unified framework for conducting backdoor attacks using natural triggers, extending beyond GAN-based synthesis networks. To prove this, we consider Stable Diffusion \cite{stable_diffusion}, the recently popular text-to-image generation model. The rest of this section is organized as follows. First, we briefly introduce Stable Diffusion. Subsequently, we explain the applicability of the proposed method to Stable Diffusion. Finally, we introduce the experimental setup and present the experimental results.

\subsection{Preliminary: Stable Diffusion}

Stable Diffusion is based on diffusion models \cite{ddpm}. In the training process of diffusion models, noise is progressively added to a training sample $\mathbf{x}_0$ through a $T$-step forward process, producing a series of noisy samples $\mathbf{x}_1, \mathbf{x}_2, ..., \mathbf{x}_T$. As $T$ grows sufficiently large, $\textbf{x}_T$ becomes Gaussian noise eventually. Given the noisy sample $\mathbf{x}_t$ and the step $t$, the training objective is to predict the added noise at step $t$. During the inference stage, $\textbf{x}_T$ is sampled from a Gaussian distribution, followed by the reverse process of multiple denoising steps to generate an image. Stable Diffusion's improvements over diffusion models are primarily embodied in two aspects. Firstly, the forward and reverse process of Stable Diffusion occur in a compressed space, achieving better efficiency. Secondly, Stable Diffusion incorporates a conditioning mechanism, thereby realizing enhanced controllability in image generation.

\begin{figure}[htbp]
%是可选项 h表示的是here在这里插入，t表示的是在页面的顶部插入
\centering
\includegraphics[scale=0.32]{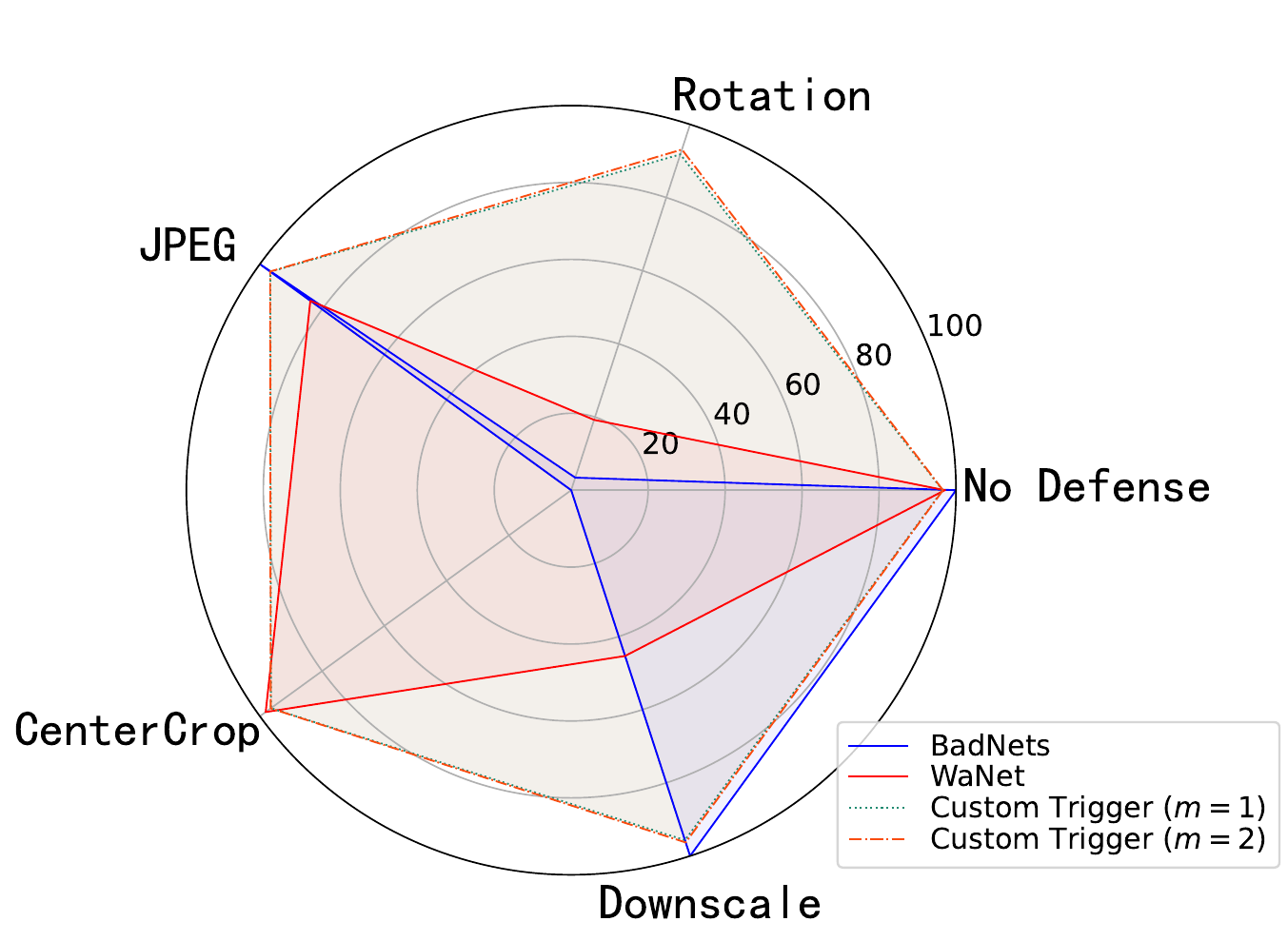}
\caption{The ASR of different attacks under various transformation-based defenses.}
\label{figure_lidar}
\end{figure}

\subsection{Trigger Customization Using Stable Diffusion}
Stable Diffusion generates images conditioned on given text prompts, implying that the manipulation of semantic features can be conducted in the text embedding space, akin to $\mathcal{W}$ space of StyleGAN. Consequently, the embedding of a natural trigger can be carried out within the text embedding space. Due to the multiple denoising steps involved in the image generation process of Stable Diffusion, the back propagation for optimization-based trigger computation demands substantial GPU memory, making such triggers less feasible. 

Compared with optimization-based trigger generation, obtaining a custom trigger for Stable Diffusion is more straightforward. We first select two prompts $p_0$ and $p_+$, where $p_0$ represents a neutral prompt (e.g., \textit{'a photo of person'}), and $p_+$ incorporates the description of the desired custom attribute(s) to $p_0$ (e.g., \textit{'a photo of person, \textbf{smile}'}). Subsequently, these two prompts are individually fed into the text encoder $E$ of Stable Diffusion to obtain their respective text embedding. The editing direction for the custom attribute $attr$ (e.g., $smile$) is then determined as the difference between these two embeddings:

\begin{equation}
  attr = E(p_+) - E(p_0).
\end{equation}

The editing direction is then scaled by the factor $\beta$ to obtain the trigger $t=\beta \cdot attr$. The attacker incorporates the trigger $t$ into the text embedding of a benign prompt $p$ to produce the poisoned embedding: $w'=w+t$, where $w=E(p)$ is the benign embedding. And poisoned samples can be generated conditioned on such poisoned embeddings. It is noteworthy that each denoising step is conditioned on the text embedding. If the trigger is embedded in all steps, the disparity between images generated with and without the trigger may not be limited to the attacker-specified attribute. For instance, the attacker intends to designate smile as the trigger, but the identity of the face also changes after the trigger embedding. This phenomenon can be attributed to the incompletely disentangled nature of Stable Diffusion, as also observed in recent works \cite{disentanglement}. To address this problem, we opt to introduce the trigger only in later denoising steps. The denoising process initiates from step $T$ and ends at step $0$ (i.e., $T \rightarrow T-1 \rightarrow \ldots \rightarrow 0$), whereas the trigger addition starts from step $T'$ and ends at step $0$ (i.e., $T' \rightarrow T'-1 \rightarrow \ldots \rightarrow 0$), where $T'<T$.

\begin{table}[htbp]
\caption{Attack Resistance against Various Transformation-Based Defenses}
\label{table_sd}
\begin{tabular}{ccccc}
\hline
Defense                     & Attack                 & BA $\uparrow$ & ASR $\uparrow$ & ABA $\uparrow$ \\ \hline
\multirow{4}{*}{No defense} & BadNets \cite{badnets}               & 99.98         & 100.0          & 100.0          \\
                            & WaNet \cite{wanet}                 & 99.97         & 97.09          & 96.78          \\
                            & Custom Trigger ($m=1$) & 99.86         & 96.53          & 97.40          \\
                            & Custom Trigger ($m=2$) & 99.83         & 96.62          & 98.44          \\ \hline
\multirow{4}{*}{Rotation}   & BadNets \cite{badnets}               & 99.04         & 3.43           & 96.57          \\
                            & WaNet \cite{wanet}                 & 94.46         & 19.23          & 99.38          \\
                            & Custom Trigger ($m=1$) & 95.68         & 91.80          & 97.40          \\
                            & Custom Trigger ($m=2$) & 97.49         & 93.13          & 97.51          \\ \hline
\multirow{4}{*}{JPEG}       & BadNets \cite{badnets}               & 99.83         & 100.0          & 99.58          \\
                            & WaNet \cite{wanet}                 & 99.88         & 83.78          & 97.51          \\
                            & Custom Trigger ($m=1$) & 99.75         & 96.64          & 97.61          \\
                            & Custom Trigger ($m=2$) & 99.65         & 96.72          & 98.13          \\ \hline
\multirow{4}{*}{CenterCrop} & BadNets \cite{badnets}               & 99.99         & 0.0            & 100.0          \\
                            & WaNet \cite{wanet}                 & 99.96         & 98.13          & 97.92          \\
                            & Custom Trigger ($m=1$) & 99.77         & 96.42          & 97.09          \\
                            & Custom Trigger ($m=2$) & 99.83         & 96.51          & 98.44          \\ \hline
\multirow{4}{*}{Downscale}  & BadNets \cite{badnets}               & 99.88         & 100.0          & 99.90          \\
                            & WaNet \cite{wanet}                 & 99.77         & 45.32          & 99.58          \\
                            & Custom Trigger ($m=1$) & 99.77         & 95.69          & 97.92          \\
                            & Custom Trigger ($m=2$) & 99.77         & 96.30          & 98.54          \\ \hline
\end{tabular}
\end{table}

To obtain the custom trigger editing multiple attributes (i.e., $m > 1$), a straightforward approach is to include all attacker-specified attributes in $p_+$. For instance, if the target attributes are age and smile, then $p_+$ can be \textit{'a photo of person, \textbf{smile, child}'}. However, due to the incompletely disentangled nature of Stable Diffusion mentioned earlier, such a trigger may introduce undesirable alterations to poisoned images, even when restricting the trigger addition steps. To mitigate this issue, we adopt a strategy of adding different single-attribute triggers in different steps to achieve multiple-attributes editing. For the smile $\&$ age case mentioned above, the attacker can introduce the smile trigger from step $T'$ to step $T''$ (i.e., $T' \rightarrow T'-1 \rightarrow \ldots \rightarrow T''$), and include the age trigger from step $T''$ to step $0$ (i.e., $T'' \rightarrow T''-1 \rightarrow \ldots \rightarrow 0$), where $T''<T'<T$.

Poisoned images generated by Stable Diffusion are depicted in Figure \ref{figure_custom_sample_sd}. For the custom trigger editing only smile (i.e., $m=1$), the trigger $t=\beta_1 \cdot smile$ is embedded from step $T'=0.7T$ to step $0$.For the custom trigger editing both smile and age, the trigger $t=\beta_1 \cdot smile$ is embedded from step $T'=0.8T$ to step $T''=0.4T$, and the trigger $t=\beta_2 \cdot age$ is embedded from step $T''=0.4T$ to step $0$. Overall, the poisoned samples appear natural. Note that in some generated faces, the teeth regions seem abnormal, which is an inherent limitation within Stable Diffusion.

\subsection{Evaluation}

Next, we introduce the experimental setup. Most settings are the same as those described in Section \ref{section4.1}, except for the fake images collection. Given the consideration of Stable Diffusion, the original fake images (i.e., fake images before injecting poisoned ones) need to contain images generated by Stable Diffusion. Concretely, the original fake images consist of 10000 PGGAN generated images, 10000 StyleGAN generated images and 10000 Stable Diffusion (version 1.4) generated images. For Stable Diffusion, we utilize the prompts provided in Papa et al.~\cite{sd_detect} to generate images. The scale factor $\beta_1$ for single attribute editing (i.e., $m=1$) is set to 1.0, and both scale factors $\beta_1$ and $\beta_2$ for multiple attributes editing (i.e., $m=2$) are set to 1.0. For comparison, BadNets and Wanet are selected as representative methods for visible and invisible backdoor attacks, respectively. The poisoning rate is set to 3.99$\%$ for all attacks.

To further demonstrate the robustness of the proposed method, we employ four challenging types of transformation-based defenses: 15 degrees rotation, JPEG compression, cropping the central 250$\times$250 region and resizing to 300$\times$300, and downscaling the image to 90$\%$ of its original size then upscaling back. The results are presented in Table \ref{table_sd}. Although the attack success rate (ASR) of the proposed method is slightly lower than baseline attacks under no defense circumstances, the proposed method is more robust against various transformation-based defenses. For visible backdoor attacks that add patches in the corner of images, the patches can be easily removed by rotation or cropping. For invisible backdoor attacks which utilize hidden patterns in pixel space, such as specific wrapping mode in WaNet, to activate backdoor, these patterns are sensitive to compression and image quality reduction. In contrast, the proposed method utilizes semantic features to trigger the backdoor, showcasing superior resistance against various transformation-based defenses. The ASR of different attacks under various conditions is depicted in Figure \ref{figure_lidar}, clearly illustrating the superior performance of the proposed method.

\section{Conclusion}
In this paper, we evaluate the robustness of face forgery detection models and backdoor defenses when confronted with natural backdoor triggers. In order to achieve this goal, we propose a novel backdoor attack by embedding the natural triggers in the latent space. We provide two ways to obtain the latent space trigger, namely the optimization-based way and the custom way. Natural semantic features created by the trigger are utilized to activate the backdoor. Furthermore, to thoroughly evaluate the detection models towards the latest AIGC, we utilize both state-of-the-art StyleGAN and Stable Diffusion for trigger generation. The experimental results show that our method is stealthier and more robust than the digital space backdoor attacks, while achieving comparable attack performance. The attack is implemented against the face forgery detection task, revealing its vulnerability to backdoor attacks. In the future, we will explore more effective defense methods to secure face forgery detection systems.

\noindent
\textbf{Ethical Statement.} Our research objective is to steer technology towards ethical applications. By exploring these attacks, we aim to unveil vulnerabilities essential for enhancing defense mechanisms, thus contributing to the development of more robust tools. Moreover, we emphasize our dedication to face privacy. The synthetic facial images showcased in our study strictly adhere to ethical standards governing the use of public data. Our research strives to balance the imperative of uncovering vulnerabilities with a steadfast commitment to privacy.

\bibliographystyle{ACM-Reference-Format}
\bibliography{sample-base}


\begin{thebibliography}{61}

%%% the Web address.

\ifx \showCODEN    \undefined \def \showCODEN     #1{\unskip}     \fi
\ifx \showDOI      \undefined \def \showDOI       #1{#1}\fi
\ifx \showISBNx    \undefined \def \showISBNx     #1{\unskip}     \fi
\ifx \showISBNxiii \undefined \def \showISBNxiii  #1{\unskip}     \fi
\ifx \showISSN     \undefined \def \showISSN      #1{\unskip}     \fi
\ifx \showLCCN     \undefined \def \showLCCN      #1{\unskip}     \fi
\ifx \shownote     \undefined \def \shownote      #1{#1}          \fi
\ifx \showarticletitle \undefined \def \showarticletitle #1{#1}   \fi
\ifx \showURL      \undefined \def \showURL       {\relax}        \fi
% invisible to TeX
\providecommand\bibfield[2]{#2}
\providecommand\bibinfo[2]{#2}
\providecommand\natexlab[1]{#1}
\providecommand\showeprint[2][]{arXiv:#2}

\bibitem[Afchar et~al\mbox{.}(2018)]%
        {afchar2018mesonet}
\bibfield{author}{\bibinfo{person}{Darius Afchar}, \bibinfo{person}{Vincent Nozick}, \bibinfo{person}{Junichi Yamagishi}, {and} \bibinfo{person}{Isao Echizen}.} \bibinfo{year}{2018}\natexlab{}.
\newblock \showarticletitle{Mesonet: a compact facial video forgery detection network}. In \bibinfo{booktitle}{\emph{2018 IEEE international workshop on information forensics and security (WIFS)}}. IEEE, \bibinfo{pages}{1--7}.
\newblock


\bibitem[Ayoubi(2021)]%
        {facelib}
\bibfield{author}{\bibinfo{person}{Sajjad Ayoubi}.} \bibinfo{year}{2021}\natexlab{}.
\newblock \bibinfo{title}{FaceLib}.
\newblock \bibinfo{howpublished}{\url{https://github.com/sajjjadayobi/FaceLib}}.
\newblock
\newblock
\shownote{Used for face detection, facial expression, AgeGender estimation and recognition with PyTorch.}.


\bibitem[Bagdasaryan et~al\mbox{.}(2020)]%
        {how_to_backdoor}
\bibfield{author}{\bibinfo{person}{Eugene Bagdasaryan}, \bibinfo{person}{Andreas Veit}, \bibinfo{person}{Yiqing Hua}, \bibinfo{person}{Deborah Estrin}, {and} \bibinfo{person}{Vitaly Shmatikov}.} \bibinfo{year}{2020}\natexlab{}.
\newblock \showarticletitle{How To Backdoor Federated Learning}. In \bibinfo{booktitle}{\emph{The 23rd International Conference on Artificial Intelligence and Statistics, {AISTATS} 2020, 26-28 August 2020, Online [Palermo, Sicily, Italy]}} \emph{(\bibinfo{series}{Proceedings of Machine Learning Research}, Vol.~\bibinfo{volume}{108})}. \bibinfo{publisher}{{PMLR}}, \bibinfo{pages}{2938--2948}.
\newblock


\bibitem[Barni et~al\mbox{.}(2019)]%
        {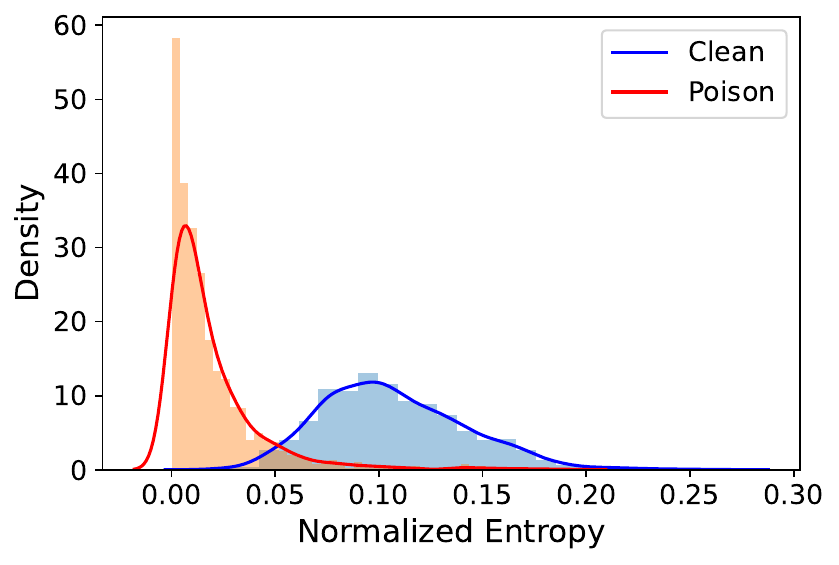}
\bibfield{author}{\bibinfo{person}{Mauro Barni}, \bibinfo{person}{Kassem Kallas}, {and} \bibinfo{person}{Benedetta Tondi}.} \bibinfo{year}{2019}\natexlab{}.
\newblock \showarticletitle{A New Backdoor Attack in {CNNS} by Training Set Corruption Without Label Poisoning}. In \bibinfo{booktitle}{\emph{2019 {IEEE} International Conference on Image Processing, {ICIP} 2019, Taipei, Taiwan, September 22-25, 2019}}. \bibinfo{publisher}{{IEEE}}, \bibinfo{pages}{101--105}.
\newblock


\bibitem[Cao et~al\mbox{.}(2022)]%
        {cao2022end}
\bibfield{author}{\bibinfo{person}{Junyi Cao}, \bibinfo{person}{Chao Ma}, \bibinfo{person}{Taiping Yao}, \bibinfo{person}{Shen Chen}, \bibinfo{person}{Shouhong Ding}, {and} \bibinfo{person}{Xiaokang Yang}.} \bibinfo{year}{2022}\natexlab{}.
\newblock \showarticletitle{End-to-end reconstruction-classification learning for face forgery detection}. In \bibinfo{booktitle}{\emph{Proceedings of the IEEE/CVF Conference on Computer Vision and Pattern Recognition}}. \bibinfo{pages}{4113--4122}.
\newblock


\bibitem[Cao and Gong(2021)]%
        {security}
\bibfield{author}{\bibinfo{person}{Xiaoyu Cao} {and} \bibinfo{person}{Neil~Zhenqiang Gong}.} \bibinfo{year}{2021}\natexlab{}.
\newblock \showarticletitle{Understanding the Security of Deepfake Detection}.
\newblock \bibinfo{journal}{\emph{CoRR}}  \bibinfo{volume}{abs/2107.02045} (\bibinfo{year}{2021}).
\newblock
\showeprint[arXiv]{2107.02045}
\urldef\tempurl%
\url{https://arxiv.org/abs/2107.02045}
\showURL{%
\tempurl}


\bibitem[Chen et~al\mbox{.}(2019)]%
        {deepinspect}
\bibfield{author}{\bibinfo{person}{Huili Chen}, \bibinfo{person}{Cheng Fu}, \bibinfo{person}{Jishen Zhao}, {and} \bibinfo{person}{Farinaz Koushanfar}.} \bibinfo{year}{2019}\natexlab{}.
\newblock \showarticletitle{DeepInspect: {A} Black-box Trojan Detection and Mitigation Framework for Deep Neural Networks}. In \bibinfo{booktitle}{\emph{Proceedings of the Twenty-Eighth International Joint Conference on Artificial Intelligence, {IJCAI} 2019, Macao, China, August 10-16, 2019}}. \bibinfo{publisher}{ijcai.org}, \bibinfo{pages}{4658--4664}.
\newblock


\bibitem[Chen et~al\mbox{.}(2017)]%
        {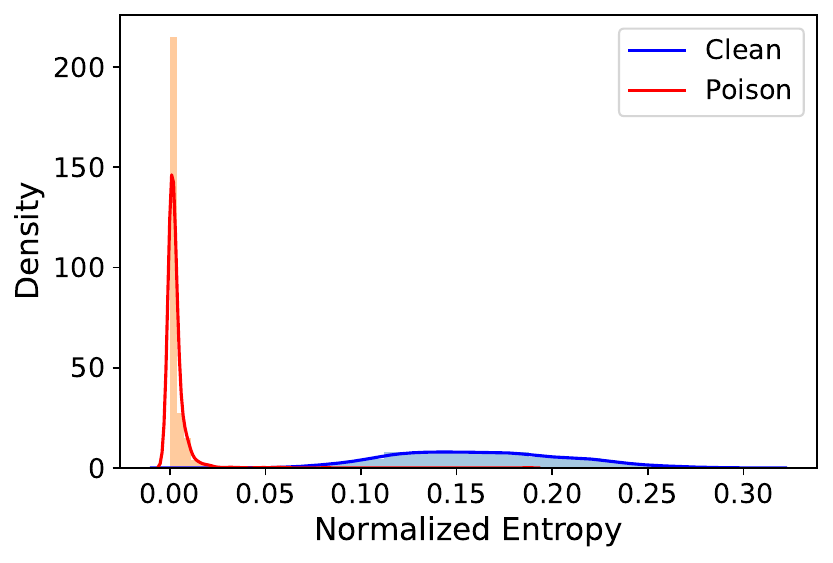}
\bibfield{author}{\bibinfo{person}{Xinyun Chen}, \bibinfo{person}{Chang Liu}, \bibinfo{person}{Bo Li}, \bibinfo{person}{Kimberly Lu}, {and} \bibinfo{person}{Dawn Song}.} \bibinfo{year}{2017}\natexlab{}.
\newblock \showarticletitle{Targeted Backdoor Attacks on Deep Learning Systems Using Data Poisoning}.
\newblock \bibinfo{journal}{\emph{CoRR}}  \bibinfo{volume}{abs/1712.05526} (\bibinfo{year}{2017}).
\newblock
\showeprint[arXiv]{1712.05526}


\bibitem[Chollet(2017)]%
        {xception}
\bibfield{author}{\bibinfo{person}{Fran{\c{c}}ois Chollet}.} \bibinfo{year}{2017}\natexlab{}.
\newblock \showarticletitle{Xception: Deep Learning with Depthwise Separable Convolutions}. In \bibinfo{booktitle}{\emph{2017 {IEEE} Conference on Computer Vision and Pattern Recognition, {CVPR} 2017, Honolulu, HI, USA, July 21-26, 2017}}. \bibinfo{publisher}{{IEEE} Computer Society}, \bibinfo{pages}{1800--1807}.
\newblock


\bibitem[Dang et~al\mbox{.}(2020)]%
        {dffd}
\bibfield{author}{\bibinfo{person}{Hao Dang}, \bibinfo{person}{Feng Liu}, \bibinfo{person}{Joel Stehouwer}, \bibinfo{person}{Xiaoming Liu}, {and} \bibinfo{person}{Anil~K. Jain}.} \bibinfo{year}{2020}\natexlab{}.
\newblock \showarticletitle{On the Detection of Digital Face Manipulation}. In \bibinfo{booktitle}{\emph{2020 {IEEE/CVF} Conference on Computer Vision and Pattern Recognition, {CVPR} 2020, Seattle, WA, USA, June 13-19, 2020}}. \bibinfo{publisher}{Computer Vision Foundation / {IEEE}}, \bibinfo{pages}{5780--5789}.
\newblock


\bibitem[Durall et~al\mbox{.}(2020)]%
        {durall2020watch}
\bibfield{author}{\bibinfo{person}{Ricard Durall}, \bibinfo{person}{Margret Keuper}, {and} \bibinfo{person}{Janis Keuper}.} \bibinfo{year}{2020}\natexlab{}.
\newblock \showarticletitle{Watch your up-convolution: Cnn based generative deep neural networks are failing to reproduce spectral distributions}. In \bibinfo{booktitle}{\emph{Proceedings of the IEEE/CVF conference on computer vision and pattern recognition}}. \bibinfo{pages}{7890--7899}.
\newblock


\bibitem[Gandhi and Jain(2020)]%
        {adv_ptb}
\bibfield{author}{\bibinfo{person}{Apurva Gandhi} {and} \bibinfo{person}{Shomik Jain}.} \bibinfo{year}{2020}\natexlab{}.
\newblock \showarticletitle{Adversarial Perturbations Fool Deepfake Detectors}. In \bibinfo{booktitle}{\emph{2020 International Joint Conference on Neural Networks, {IJCNN} 2020, Glasgow, United Kingdom, July 19-24, 2020}}. \bibinfo{publisher}{{IEEE}}, \bibinfo{pages}{1--8}.
\newblock


\bibitem[Gao et~al\mbox{.}(2019)]%
        {strip}
\bibfield{author}{\bibinfo{person}{Yansong Gao}, \bibinfo{person}{Chang Xu}, \bibinfo{person}{Derui Wang}, \bibinfo{person}{Shiping Chen}, \bibinfo{person}{Damith~Chinthana Ranasinghe}, {and} \bibinfo{person}{Surya Nepal}.} \bibinfo{year}{2019}\natexlab{}.
\newblock \showarticletitle{{STRIP:} a defence against trojan attacks on deep neural networks}. In \bibinfo{booktitle}{\emph{Proceedings of the 35th Annual Computer Security Applications Conference, {ACSAC} 2019, San Juan, PR, USA, December 09-13, 2019}}. \bibinfo{publisher}{{ACM}}, \bibinfo{pages}{113--125}.
\newblock


\bibitem[Goodfellow et~al\mbox{.}(2014)]%
        {gan}
\bibfield{author}{\bibinfo{person}{Ian~J. Goodfellow}, \bibinfo{person}{Jean Pouget{-}Abadie}, \bibinfo{person}{Mehdi Mirza}, \bibinfo{person}{Bing Xu}, \bibinfo{person}{David Warde{-}Farley}, \bibinfo{person}{Sherjil Ozair}, \bibinfo{person}{Aaron~C. Courville}, {and} \bibinfo{person}{Yoshua Bengio}.} \bibinfo{year}{2014}\natexlab{}.
\newblock \showarticletitle{Generative Adversarial Nets}. In \bibinfo{booktitle}{\emph{Advances in Neural Information Processing Systems 27: Annual Conference on Neural Information Processing Systems 2014, December 8-13 2014, Montreal, Quebec, Canada}}. \bibinfo{pages}{2672--2680}.
\newblock


\bibitem[Gu et~al\mbox{.}(2019)]%
        {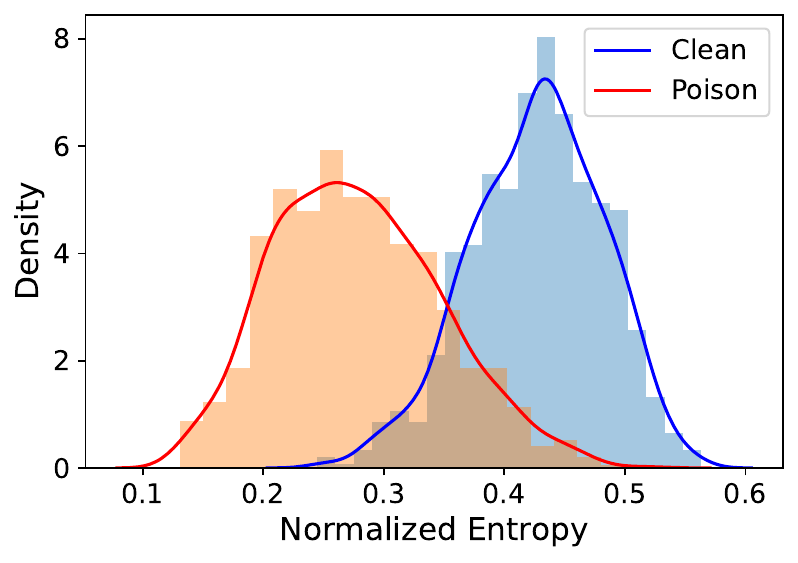}
\bibfield{author}{\bibinfo{person}{Tianyu Gu}, \bibinfo{person}{Kang Liu}, \bibinfo{person}{Brendan Dolan-Gavitt}, {and} \bibinfo{person}{Siddharth Garg}.} \bibinfo{year}{2019}\natexlab{}.
\newblock \showarticletitle{BadNets: Evaluating Backdooring Attacks on Deep Neural Networks}.
\newblock \bibinfo{journal}{\emph{IEEE Access}}  \bibinfo{volume}{7} (\bibinfo{year}{2019}), \bibinfo{pages}{47230--47244}.
\newblock


\bibitem[He et~al\mbox{.}(2016)]%
        {resnet}
\bibfield{author}{\bibinfo{person}{Kaiming He}, \bibinfo{person}{Xiangyu Zhang}, \bibinfo{person}{Shaoqing Ren}, {and} \bibinfo{person}{Jian Sun}.} \bibinfo{year}{2016}\natexlab{}.
\newblock \showarticletitle{Deep Residual Learning for Image Recognition}. In \bibinfo{booktitle}{\emph{2016 {IEEE} Conference on Computer Vision and Pattern Recognition, {CVPR} 2016, Las Vegas, NV, USA, June 27-30, 2016}}. \bibinfo{publisher}{{IEEE} Computer Society}, \bibinfo{pages}{770--778}.
\newblock


\bibitem[Ho et~al\mbox{.}(2020)]%
        {ddpm}
\bibfield{author}{\bibinfo{person}{Jonathan Ho}, \bibinfo{person}{Ajay Jain}, {and} \bibinfo{person}{Pieter Abbeel}.} \bibinfo{year}{2020}\natexlab{}.
\newblock \showarticletitle{Denoising Diffusion Probabilistic Models}. In \bibinfo{booktitle}{\emph{Advances in Neural Information Processing Systems 33: Annual Conference on Neural Information Processing Systems 2020, NeurIPS 2020, December 6-12, 2020, virtual}}.
\newblock


\bibitem[Hussain et~al\mbox{.}(2021)]%
        {adv_df}
\bibfield{author}{\bibinfo{person}{Shehzeen Hussain}, \bibinfo{person}{Paarth Neekhara}, \bibinfo{person}{Malhar Jere}, \bibinfo{person}{Farinaz Koushanfar}, {and} \bibinfo{person}{Julian~J. McAuley}.} \bibinfo{year}{2021}\natexlab{}.
\newblock \showarticletitle{Adversarial Deepfakes: Evaluating Vulnerability of Deepfake Detectors to Adversarial Examples}. In \bibinfo{booktitle}{\emph{{IEEE} Winter Conference on Applications of Computer Vision, {WACV} 2021, Waikoloa, HI, USA, January 3-8, 2021}}. \bibinfo{publisher}{{IEEE}}, \bibinfo{pages}{3347--3356}.
\newblock


\bibitem[Jia et~al\mbox{.}(2022)]%
        {exp_fre}
\bibfield{author}{\bibinfo{person}{Shuai Jia}, \bibinfo{person}{Chao Ma}, \bibinfo{person}{Taiping Yao}, \bibinfo{person}{Bangjie Yin}, \bibinfo{person}{Shouhong Ding}, {and} \bibinfo{person}{Xiaokang Yang}.} \bibinfo{year}{2022}\natexlab{}.
\newblock \showarticletitle{Exploring Frequency Adversarial Attacks for Face Forgery Detection}. In \bibinfo{booktitle}{\emph{{IEEE/CVF} Conference on Computer Vision and Pattern Recognition, {CVPR} 2022, New Orleans, LA, USA, June 18-24, 2022}}. \bibinfo{publisher}{{IEEE}}, \bibinfo{pages}{4093--4102}.
\newblock


\bibitem[Karras et~al\mbox{.}(2018)]%
        {pggan}
\bibfield{author}{\bibinfo{person}{Tero Karras}, \bibinfo{person}{Timo Aila}, \bibinfo{person}{Samuli Laine}, {and} \bibinfo{person}{Jaakko Lehtinen}.} \bibinfo{year}{2018}\natexlab{}.
\newblock \showarticletitle{Progressive Growing of GANs for Improved Quality, Stability, and Variation}. In \bibinfo{booktitle}{\emph{6th International Conference on Learning Representations, {ICLR} 2018, Vancouver, BC, Canada, April 30 - May 3, 2018, Conference Track Proceedings}}. \bibinfo{publisher}{OpenReview.net}.
\newblock


\bibitem[Karras et~al\mbox{.}(2019)]%
        {stylegan}
\bibfield{author}{\bibinfo{person}{Tero Karras}, \bibinfo{person}{Samuli Laine}, {and} \bibinfo{person}{Timo Aila}.} \bibinfo{year}{2019}\natexlab{}.
\newblock \showarticletitle{A Style-Based Generator Architecture for Generative Adversarial Networks}. In \bibinfo{booktitle}{\emph{{IEEE} Conference on Computer Vision and Pattern Recognition, {CVPR} 2019, Long Beach, CA, USA, June 16-20, 2019}}. \bibinfo{publisher}{Computer Vision Foundation / {IEEE}}, \bibinfo{pages}{4401--4410}.
\newblock


\bibitem[Kingma and Ba(2015)]%
        {adam}
\bibfield{author}{\bibinfo{person}{Diederik~P. Kingma} {and} \bibinfo{person}{Jimmy Ba}.} \bibinfo{year}{2015}\natexlab{}.
\newblock \showarticletitle{Adam: {A} Method for Stochastic Optimization}. In \bibinfo{booktitle}{\emph{3rd International Conference on Learning Representations, {ICLR} 2015, San Diego, CA, USA, May 7-9, 2015, Conference Track Proceedings}}.
\newblock


\bibitem[Kristanto et~al\mbox{.}(2022)]%
        {latent_bd}
\bibfield{author}{\bibinfo{person}{Adrian Kristanto}, \bibinfo{person}{Shuo Wang}, {and} \bibinfo{person}{Carsten Rudolph}.} \bibinfo{year}{2022}\natexlab{}.
\newblock \showarticletitle{Latent Space-Based Backdoor Attacks Against Deep Neural Networks}. In \bibinfo{booktitle}{\emph{International Joint Conference on Neural Networks, {IJCNN} 2022, Padua, Italy, July 18-23, 2022}}. \bibinfo{publisher}{{IEEE}}, \bibinfo{pages}{1--10}.
\newblock


\bibitem[Li et~al\mbox{.}(2021c)]%
        {exp_adv}
\bibfield{author}{\bibinfo{person}{Dongze Li}, \bibinfo{person}{Wei Wang}, \bibinfo{person}{Hongxing Fan}, {and} \bibinfo{person}{Jing Dong}.} \bibinfo{year}{2021}\natexlab{c}.
\newblock \showarticletitle{Exploring Adversarial Fake Images on Face Manifold}. In \bibinfo{booktitle}{\emph{{IEEE} Conference on Computer Vision and Pattern Recognition, {CVPR} 2021, virtual, June 19-25, 2021}}. \bibinfo{publisher}{Computer Vision Foundation / {IEEE}}, \bibinfo{pages}{5789--5798}.
\newblock


\bibitem[Li et~al\mbox{.}(2021a)]%
        {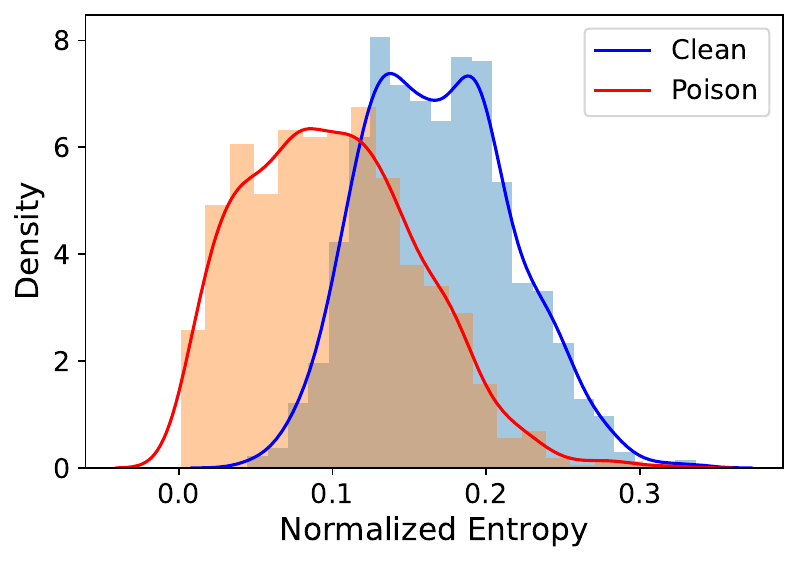}
\bibfield{author}{\bibinfo{person}{Yuezun Li}, \bibinfo{person}{Yiming Li}, \bibinfo{person}{Baoyuan Wu}, \bibinfo{person}{Longkang Li}, \bibinfo{person}{Ran He}, {and} \bibinfo{person}{Siwei Lyu}.} \bibinfo{year}{2021}\natexlab{a}.
\newblock \showarticletitle{Invisible Backdoor Attack with Sample-Specific Triggers}. In \bibinfo{booktitle}{\emph{2021 {IEEE/CVF} International Conference on Computer Vision, {ICCV} 2021, Montreal, QC, Canada, October 10-17, 2021}}. \bibinfo{publisher}{{IEEE}}, \bibinfo{pages}{16443--16452}.
\newblock


\bibitem[Li et~al\mbox{.}(2021b)]%
        {nad}
\bibfield{author}{\bibinfo{person}{Yige Li}, \bibinfo{person}{Xixiang Lyu}, \bibinfo{person}{Nodens Koren}, \bibinfo{person}{Lingjuan Lyu}, \bibinfo{person}{Bo Li}, {and} \bibinfo{person}{Xingjun Ma}.} \bibinfo{year}{2021}\natexlab{b}.
\newblock \showarticletitle{Neural Attention Distillation: Erasing Backdoor Triggers from Deep Neural Networks}. In \bibinfo{booktitle}{\emph{9th International Conference on Learning Representations, {ICLR} 2021, Virtual Event, Austria, May 3-7, 2021}}. \bibinfo{publisher}{OpenReview.net}.
\newblock


\bibitem[Li et~al\mbox{.}(2020)]%
        {rethinking}
\bibfield{author}{\bibinfo{person}{Yiming Li}, \bibinfo{person}{Tongqing Zhai}, \bibinfo{person}{Baoyuan Wu}, \bibinfo{person}{Yong Jiang}, \bibinfo{person}{Zhifeng Li}, {and} \bibinfo{person}{Shutao Xia}.} \bibinfo{year}{2020}\natexlab{}.
\newblock \showarticletitle{Rethinking the Trigger of Backdoor Attack}.
\newblock \bibinfo{journal}{\emph{CoRR}}  \bibinfo{volume}{abs/2004.04692} (\bibinfo{year}{2020}).
\newblock
\showeprint[arXiv]{2004.04692}
\urldef\tempurl%
\url{https://arxiv.org/abs/2004.04692}
\showURL{%
\tempurl}


\bibitem[Liang et~al\mbox{.}(2022)]%
        {liang2022exploring}
\bibfield{author}{\bibinfo{person}{Jiahao Liang}, \bibinfo{person}{Huafeng Shi}, {and} \bibinfo{person}{Weihong Deng}.} \bibinfo{year}{2022}\natexlab{}.
\newblock \showarticletitle{Exploring disentangled content information for face forgery detection}. In \bibinfo{booktitle}{\emph{European Conference on Computer Vision}}. Springer, \bibinfo{pages}{128--145}.
\newblock


\bibitem[Lin et~al\mbox{.}(2020)]%
        {composite}
\bibfield{author}{\bibinfo{person}{Junyu Lin}, \bibinfo{person}{Lei Xu}, \bibinfo{person}{Yingqi Liu}, {and} \bibinfo{person}{Xiangyu Zhang}.} \bibinfo{year}{2020}\natexlab{}.
\newblock \showarticletitle{Composite Backdoor Attack for Deep Neural Network by Mixing Existing Benign Features}. In \bibinfo{booktitle}{\emph{{CCS} '20: 2020 {ACM} {SIGSAC} Conference on Computer and Communications Security, Virtual Event, USA, November 9-13, 2020}}. \bibinfo{publisher}{{ACM}}, \bibinfo{pages}{113--131}.
\newblock


\bibitem[Liu et~al\mbox{.}(2021)]%
        {liu2021spatial}
\bibfield{author}{\bibinfo{person}{Honggu Liu}, \bibinfo{person}{Xiaodan Li}, \bibinfo{person}{Wenbo Zhou}, \bibinfo{person}{Yuefeng Chen}, \bibinfo{person}{Yuan He}, \bibinfo{person}{Hui Xue}, \bibinfo{person}{Weiming Zhang}, {and} \bibinfo{person}{Nenghai Yu}.} \bibinfo{year}{2021}\natexlab{}.
\newblock \showarticletitle{Spatial-phase shallow learning: rethinking face forgery detection in frequency domain}. In \bibinfo{booktitle}{\emph{Proceedings of the IEEE/CVF conference on computer vision and pattern recognition}}. \bibinfo{pages}{772--781}.
\newblock


\bibitem[Liu et~al\mbox{.}(2018)]%
        {fine_prun}
\bibfield{author}{\bibinfo{person}{Kang Liu}, \bibinfo{person}{Brendan Dolan{-}Gavitt}, {and} \bibinfo{person}{Siddharth Garg}.} \bibinfo{year}{2018}\natexlab{}.
\newblock \showarticletitle{Fine-Pruning: Defending Against Backdooring Attacks on Deep Neural Networks}. In \bibinfo{booktitle}{\emph{Research in Attacks, Intrusions, and Defenses - 21st International Symposium, {RAID} 2018, Heraklion, Crete, Greece, September 10-12, 2018, Proceedings}} \emph{(\bibinfo{series}{Lecture Notes in Computer Science}, Vol.~\bibinfo{volume}{11050})}. \bibinfo{publisher}{Springer}, \bibinfo{pages}{273--294}.
\newblock


\bibitem[Liu et~al\mbox{.}(2020)]%
        {refool}
\bibfield{author}{\bibinfo{person}{Yunfei Liu}, \bibinfo{person}{Xingjun Ma}, \bibinfo{person}{James Bailey}, {and} \bibinfo{person}{Feng Lu}.} \bibinfo{year}{2020}\natexlab{}.
\newblock \showarticletitle{Reflection Backdoor: {A} Natural Backdoor Attack on Deep Neural Networks}. In \bibinfo{booktitle}{\emph{Computer Vision - {ECCV} 2020 - 16th European Conference, Glasgow, UK, August 23-28, 2020, Proceedings, Part {X}}} \emph{(\bibinfo{series}{Lecture Notes in Computer Science}, Vol.~\bibinfo{volume}{12355})}. \bibinfo{publisher}{Springer}, \bibinfo{pages}{182--199}.
\newblock


\bibitem[Liu et~al\mbox{.}(2015)]%
        {celeba}
\bibfield{author}{\bibinfo{person}{Ziwei Liu}, \bibinfo{person}{Ping Luo}, \bibinfo{person}{Xiaogang Wang}, {and} \bibinfo{person}{Xiaoou Tang}.} \bibinfo{year}{2015}\natexlab{}.
\newblock \showarticletitle{Deep Learning Face Attributes in the Wild}. In \bibinfo{booktitle}{\emph{Proceedings of International Conference on Computer Vision (ICCV)}}.
\newblock


\bibitem[Luo et~al\mbox{.}(2021)]%
        {luo2021generalizing}
\bibfield{author}{\bibinfo{person}{Yuchen Luo}, \bibinfo{person}{Yong Zhang}, \bibinfo{person}{Junchi Yan}, {and} \bibinfo{person}{Wei Liu}.} \bibinfo{year}{2021}\natexlab{}.
\newblock \showarticletitle{Generalizing face forgery detection with high-frequency features}. In \bibinfo{booktitle}{\emph{Proceedings of the IEEE/CVF conference on computer vision and pattern recognition}}. \bibinfo{pages}{16317--16326}.
\newblock


\bibitem[Ma et~al\mbox{.}(2018)]%
        {shufflenet}
\bibfield{author}{\bibinfo{person}{Ningning Ma}, \bibinfo{person}{Xiangyu Zhang}, \bibinfo{person}{Hai{-}Tao Zheng}, {and} \bibinfo{person}{Jian Sun}.} \bibinfo{year}{2018}\natexlab{}.
\newblock \showarticletitle{ShuffleNet {V2:} Practical Guidelines for Efficient {CNN} Architecture Design}. In \bibinfo{booktitle}{\emph{Computer Vision - {ECCV} 2018 - 15th European Conference, Munich, Germany, September 8-14, 2018, Proceedings, Part {XIV}}} \emph{(\bibinfo{series}{Lecture Notes in Computer Science}, Vol.~\bibinfo{volume}{11218})}. \bibinfo{publisher}{Springer}, \bibinfo{pages}{122--138}.
\newblock


\bibitem[Moosavi{-}Dezfooli et~al\mbox{.}(2017)]%
        {Universal}
\bibfield{author}{\bibinfo{person}{Seyed{-}Mohsen Moosavi{-}Dezfooli}, \bibinfo{person}{Alhussein Fawzi}, \bibinfo{person}{Omar Fawzi}, {and} \bibinfo{person}{Pascal Frossard}.} \bibinfo{year}{2017}\natexlab{}.
\newblock \showarticletitle{Universal Adversarial Perturbations}. In \bibinfo{booktitle}{\emph{2017 {IEEE} Conference on Computer Vision and Pattern Recognition, {CVPR} 2017, Honolulu, HI, USA, July 21-26, 2017}}. \bibinfo{publisher}{{IEEE} Computer Society}, \bibinfo{pages}{86--94}.
\newblock


\bibitem[Neekhara et~al\mbox{.}(2021)]%
        {adv_thrt}
\bibfield{author}{\bibinfo{person}{Paarth Neekhara}, \bibinfo{person}{Brian Dolhansky}, \bibinfo{person}{Joanna Bitton}, {and} \bibinfo{person}{Cristian Canton{-}Ferrer}.} \bibinfo{year}{2021}\natexlab{}.
\newblock \showarticletitle{Adversarial Threats to DeepFake Detection: {A} Practical Perspective}. In \bibinfo{booktitle}{\emph{{IEEE} Conference on Computer Vision and Pattern Recognition Workshops, {CVPR} Workshops 2021, virtual, June 19-25, 2021}}. \bibinfo{publisher}{Computer Vision Foundation / {IEEE}}, \bibinfo{pages}{923--932}.
\newblock


\bibitem[Nguyen et~al\mbox{.}(2019a)]%
        {nguyen2019multi}
\bibfield{author}{\bibinfo{person}{Huy~H Nguyen}, \bibinfo{person}{Fuming Fang}, \bibinfo{person}{Junichi Yamagishi}, {and} \bibinfo{person}{Isao Echizen}.} \bibinfo{year}{2019}\natexlab{a}.
\newblock \showarticletitle{Multi-task learning for detecting and segmenting manipulated facial images and videos}. In \bibinfo{booktitle}{\emph{2019 IEEE 10th international conference on biometrics theory, applications and systems (BTAS)}}. IEEE, \bibinfo{pages}{1--8}.
\newblock


\bibitem[Nguyen et~al\mbox{.}(2019b)]%
        {nguyen2019capsule}
\bibfield{author}{\bibinfo{person}{Huy~H Nguyen}, \bibinfo{person}{Junichi Yamagishi}, {and} \bibinfo{person}{Isao Echizen}.} \bibinfo{year}{2019}\natexlab{b}.
\newblock \showarticletitle{Capsule-forensics: Using capsule networks to detect forged images and videos}. In \bibinfo{booktitle}{\emph{ICASSP 2019-2019 IEEE International Conference on Acoustics, Speech and Signal Processing (ICASSP)}}. IEEE, \bibinfo{pages}{2307--2311}.
\newblock


\bibitem[Nguyen and Tran(2020)]%
        {input_aware}
\bibfield{author}{\bibinfo{person}{Tuan~Anh Nguyen} {and} \bibinfo{person}{Anh~Tuan Tran}.} \bibinfo{year}{2020}\natexlab{}.
\newblock \showarticletitle{Input-Aware Dynamic Backdoor Attack}. In \bibinfo{booktitle}{\emph{Advances in Neural Information Processing Systems 33: Annual Conference on Neural Information Processing Systems 2020, NeurIPS 2020, December 6-12, 2020, virtual}}.
\newblock


\bibitem[Nguyen and Tran(2021)]%
        {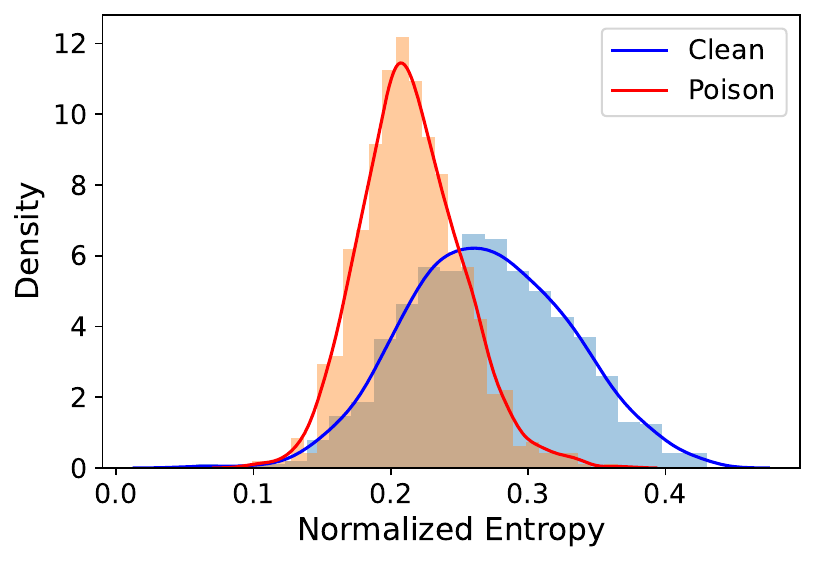}
\bibfield{author}{\bibinfo{person}{Tuan~Anh Nguyen} {and} \bibinfo{person}{Anh~Tuan Tran}.} \bibinfo{year}{2021}\natexlab{}.
\newblock \showarticletitle{WaNet - Imperceptible Warping-based Backdoor Attack}. In \bibinfo{booktitle}{\emph{9th International Conference on Learning Representations, {ICLR} 2021, Virtual Event, Austria, May 3-7, 2021}}. \bibinfo{publisher}{OpenReview.net}.
\newblock


\bibitem[Papa et~al\mbox{.}(2023)]%
        {sd_detect}
\bibfield{author}{\bibinfo{person}{Lorenzo Papa}, \bibinfo{person}{Lorenzo Faiella}, \bibinfo{person}{Luca Corvitto}, \bibinfo{person}{Luca Maiano}, {and} \bibinfo{person}{Irene Amerini}.} \bibinfo{year}{2023}\natexlab{}.
\newblock \showarticletitle{On the use of Stable Diffusion for creating realistic faces: from generation to detection}. In \bibinfo{booktitle}{\emph{11th International Workshop on Biometrics and Forensics, {IWBF} 2023, Barcelona, Spain, April 19-20, 2023}}. \bibinfo{publisher}{{IEEE}}, \bibinfo{pages}{1--6}.
\newblock


\bibitem[Peng et~al\mbox{.}(2021)]%
        {peng2021unified}
\bibfield{author}{\bibinfo{person}{Bo Peng}, \bibinfo{person}{Hongxing Fan}, \bibinfo{person}{Wei Wang}, \bibinfo{person}{Jing Dong}, {and} \bibinfo{person}{Siwei Lyu}.} \bibinfo{year}{2021}\natexlab{}.
\newblock \showarticletitle{A unified framework for high fidelity face swap and expression reenactment}.
\newblock \bibinfo{journal}{\emph{IEEE Transactions on Circuits and Systems for Video Technology}} \bibinfo{volume}{32}, \bibinfo{number}{6} (\bibinfo{year}{2021}), \bibinfo{pages}{3673--3684}.
\newblock


\bibitem[Qian et~al\mbox{.}(2020)]%
        {qian2020thinking}
\bibfield{author}{\bibinfo{person}{Yuyang Qian}, \bibinfo{person}{Guojun Yin}, \bibinfo{person}{Lu Sheng}, \bibinfo{person}{Zixuan Chen}, {and} \bibinfo{person}{Jing Shao}.} \bibinfo{year}{2020}\natexlab{}.
\newblock \showarticletitle{Thinking in frequency: Face forgery detection by mining frequency-aware clues}. In \bibinfo{booktitle}{\emph{European conference on computer vision}}. Springer, \bibinfo{pages}{86--103}.
\newblock


\bibitem[Rombach et~al\mbox{.}(2022)]%
        {stable_diffusion}
\bibfield{author}{\bibinfo{person}{Robin Rombach}, \bibinfo{person}{Andreas Blattmann}, \bibinfo{person}{Dominik Lorenz}, \bibinfo{person}{Patrick Esser}, {and} \bibinfo{person}{Bj{\"{o}}rn Ommer}.} \bibinfo{year}{2022}\natexlab{}.
\newblock \showarticletitle{High-Resolution Image Synthesis with Latent Diffusion Models}. In \bibinfo{booktitle}{\emph{{IEEE/CVF} Conference on Computer Vision and Pattern Recognition, {CVPR} 2022, New Orleans, LA, USA, June 18-24, 2022}}. \bibinfo{publisher}{{IEEE}}, \bibinfo{pages}{10674--10685}.
\newblock


\bibitem[Rossler et~al\mbox{.}(2019)]%
        {rossler2019faceforensics++}
\bibfield{author}{\bibinfo{person}{Andreas Rossler}, \bibinfo{person}{Davide Cozzolino}, \bibinfo{person}{Luisa Verdoliva}, \bibinfo{person}{Christian Riess}, \bibinfo{person}{Justus Thies}, {and} \bibinfo{person}{Matthias Nie{\ss}ner}.} \bibinfo{year}{2019}\natexlab{}.
\newblock \showarticletitle{Faceforensics++: Learning to detect manipulated facial images}. In \bibinfo{booktitle}{\emph{Proceedings of the IEEE/CVF international conference on computer vision}}. \bibinfo{pages}{1--11}.
\newblock


\bibitem[Salem et~al\mbox{.}(2022)]%
        {dynamic}
\bibfield{author}{\bibinfo{person}{Ahmed Salem}, \bibinfo{person}{Rui Wen}, \bibinfo{person}{Michael Backes}, \bibinfo{person}{Shiqing Ma}, {and} \bibinfo{person}{Yang Zhang}.} \bibinfo{year}{2022}\natexlab{}.
\newblock \showarticletitle{Dynamic Backdoor Attacks Against Machine Learning Models}. In \bibinfo{booktitle}{\emph{7th {IEEE} European Symposium on Security and Privacy, EuroS{\&}P 2022, Genoa, Italy, June 6-10, 2022}}. \bibinfo{publisher}{{IEEE}}, \bibinfo{pages}{703--718}.
\newblock


\bibitem[Sarkar et~al\mbox{.}(2022)]%
        {facehack}
\bibfield{author}{\bibinfo{person}{Esha Sarkar}, \bibinfo{person}{Hadjer Benkraouda}, \bibinfo{person}{Gopika Krishnan}, \bibinfo{person}{Homer Gamil}, {and} \bibinfo{person}{Michail Maniatakos}.} \bibinfo{year}{2022}\natexlab{}.
\newblock \showarticletitle{FaceHack: Attacking Facial Recognition Systems Using Malicious Facial Characteristics}.
\newblock \bibinfo{journal}{\emph{{IEEE} Trans. Biom. Behav. Identity Sci.}} \bibinfo{volume}{4}, \bibinfo{number}{3} (\bibinfo{year}{2022}), \bibinfo{pages}{361--372}.
\newblock


\bibitem[Selvaraju et~al\mbox{.}(2016)]%
        {gradcam}
\bibfield{author}{\bibinfo{person}{Ramprasaath~R. Selvaraju}, \bibinfo{person}{Abhishek Das}, \bibinfo{person}{Ramakrishna Vedantam}, \bibinfo{person}{Michael Cogswell}, \bibinfo{person}{Devi Parikh}, {and} \bibinfo{person}{Dhruv Batra}.} \bibinfo{year}{2016}\natexlab{}.
\newblock \showarticletitle{Grad-CAM: Visual Explanations from Deep Networks via Gradient-Based Localization}.
\newblock \bibinfo{journal}{\emph{International Journal of Computer Vision}}  \bibinfo{volume}{128} (\bibinfo{year}{2016}), \bibinfo{pages}{336--359}.
\newblock


\bibitem[Shen et~al\mbox{.}(2020)]%
        {interfacegan}
\bibfield{author}{\bibinfo{person}{Yujun Shen}, \bibinfo{person}{Jinjin Gu}, \bibinfo{person}{Xiaoou Tang}, {and} \bibinfo{person}{Bolei Zhou}.} \bibinfo{year}{2020}\natexlab{}.
\newblock \showarticletitle{Interpreting the Latent Space of GANs for Semantic Face Editing}. In \bibinfo{booktitle}{\emph{2020 {IEEE/CVF} Conference on Computer Vision and Pattern Recognition, {CVPR} 2020, Seattle, WA, USA, June 13-19, 2020}}. \bibinfo{publisher}{Computer Vision Foundation / {IEEE}}, \bibinfo{pages}{9240--9249}.
\newblock


\bibitem[Simonyan and Zisserman(2015)]%
        {vgg}
\bibfield{author}{\bibinfo{person}{Karen Simonyan} {and} \bibinfo{person}{Andrew Zisserman}.} \bibinfo{year}{2015}\natexlab{}.
\newblock \showarticletitle{Very Deep Convolutional Networks for Large-Scale Image Recognition}. In \bibinfo{booktitle}{\emph{3rd International Conference on Learning Representations, {ICLR} 2015, San Diego, CA, USA, May 7-9, 2015, Conference Track Proceedings}}.
\newblock


\bibitem[Tan and Le(2019)]%
        {efficientnet}
\bibfield{author}{\bibinfo{person}{Mingxing Tan} {and} \bibinfo{person}{Quoc~V. Le}.} \bibinfo{year}{2019}\natexlab{}.
\newblock \showarticletitle{EfficientNet: Rethinking Model Scaling for Convolutional Neural Networks}. In \bibinfo{booktitle}{\emph{Proceedings of the 36th International Conference on Machine Learning, {ICML} 2019, 9-15 June 2019, Long Beach, California, {USA}}} \emph{(\bibinfo{series}{Proceedings of Machine Learning Research}, Vol.~\bibinfo{volume}{97})}. \bibinfo{publisher}{{PMLR}}, \bibinfo{pages}{6105--6114}.
\newblock


\bibitem[Tolosana et~al\mbox{.}(2020)]%
        {tolosana2020deepfakes}
\bibfield{author}{\bibinfo{person}{Ruben Tolosana}, \bibinfo{person}{Ruben Vera-Rodriguez}, \bibinfo{person}{Julian Fierrez}, \bibinfo{person}{Aythami Morales}, {and} \bibinfo{person}{Javier Ortega-Garcia}.} \bibinfo{year}{2020}\natexlab{}.
\newblock \showarticletitle{Deepfakes and beyond: A survey of face manipulation and fake detection}.
\newblock \bibinfo{journal}{\emph{Information Fusion}}  \bibinfo{volume}{64} (\bibinfo{year}{2020}), \bibinfo{pages}{131--148}.
\newblock


\bibitem[Wang et~al\mbox{.}(2019)]%
        {neural_cleanse}
\bibfield{author}{\bibinfo{person}{Bolun Wang}, \bibinfo{person}{Yuanshun Yao}, \bibinfo{person}{Shawn Shan}, \bibinfo{person}{Huiying Li}, \bibinfo{person}{Bimal Viswanath}, \bibinfo{person}{Haitao Zheng}, {and} \bibinfo{person}{Ben~Y. Zhao}.} \bibinfo{year}{2019}\natexlab{}.
\newblock \showarticletitle{Neural Cleanse: Identifying and Mitigating Backdoor Attacks in Neural Networks}. In \bibinfo{booktitle}{\emph{2019 {IEEE} Symposium on Security and Privacy, {SP} 2019, San Francisco, CA, USA, May 19-23, 2019}}. \bibinfo{publisher}{{IEEE}}, \bibinfo{pages}{707--723}.
\newblock


\bibitem[Wang and Deng(2021)]%
        {wang2021representative}
\bibfield{author}{\bibinfo{person}{Chengrui Wang} {and} \bibinfo{person}{Weihong Deng}.} \bibinfo{year}{2021}\natexlab{}.
\newblock \showarticletitle{Representative forgery mining for fake face detection}. In \bibinfo{booktitle}{\emph{Proceedings of the IEEE/CVF conference on computer vision and pattern recognition}}. \bibinfo{pages}{14923--14932}.
\newblock


\bibitem[Wu et~al\mbox{.}(2023)]%
        {disentanglement}
\bibfield{author}{\bibinfo{person}{Qiucheng Wu}, \bibinfo{person}{Yujian Liu}, \bibinfo{person}{Handong Zhao}, \bibinfo{person}{Ajinkya Kale}, \bibinfo{person}{Trung Bui}, \bibinfo{person}{Tong Yu}, \bibinfo{person}{Zhe Lin}, \bibinfo{person}{Yang Zhang}, {and} \bibinfo{person}{Shiyu Chang}.} \bibinfo{year}{2023}\natexlab{}.
\newblock \showarticletitle{Uncovering the Disentanglement Capability in Text-to-Image Diffusion Models}. In \bibinfo{booktitle}{\emph{{IEEE/CVF} Conference on Computer Vision and Pattern Recognition, {CVPR} 2023, Vancouver, BC, Canada, June 17-24, 2023}}. \bibinfo{publisher}{{IEEE}}, \bibinfo{pages}{1900--1910}.
\newblock


\bibitem[Yang et~al\mbox{.}(2023a)]%
        {yang2023learning}
\bibfield{author}{\bibinfo{person}{Songlin Yang}, \bibinfo{person}{Wei Wang}, \bibinfo{person}{Yushi Lan}, \bibinfo{person}{Xiangyu Fan}, \bibinfo{person}{Bo Peng}, \bibinfo{person}{Lei Yang}, {and} \bibinfo{person}{Jing Dong}.} \bibinfo{year}{2023}\natexlab{a}.
\newblock \showarticletitle{Learning Dense Correspondence for NeRF-Based Face Reenactment}.
\newblock \bibinfo{journal}{\emph{arXiv preprint arXiv:2312.10422}} (\bibinfo{year}{2023}).
\newblock


\bibitem[Yang et~al\mbox{.}(2023b)]%
        {yang2023designing}
\bibfield{author}{\bibinfo{person}{Songlin Yang}, \bibinfo{person}{Wei Wang}, \bibinfo{person}{Bo Peng}, {and} \bibinfo{person}{Jing Dong}.} \bibinfo{year}{2023}\natexlab{b}.
\newblock \showarticletitle{Designing A 3d-Aware Stylenerf Encoder for Face Editing}. In \bibinfo{booktitle}{\emph{ICASSP 2023-2023 IEEE International Conference on Acoustics, Speech and Signal Processing (ICASSP)}}. IEEE, \bibinfo{pages}{1--5}.
\newblock


\bibitem[Zhang et~al\mbox{.}(2016)]%
        {mtcnn}
\bibfield{author}{\bibinfo{person}{Kaipeng Zhang}, \bibinfo{person}{Zhanpeng Zhang}, \bibinfo{person}{Zhifeng Li}, {and} \bibinfo{person}{Yu Qiao}.} \bibinfo{year}{2016}\natexlab{}.
\newblock \showarticletitle{Joint Face Detection and Alignment Using Multitask Cascaded Convolutional Networks}.
\newblock \bibinfo{journal}{\emph{IEEE Signal Processing Letters}}  \bibinfo{volume}{23} (\bibinfo{year}{2016}), \bibinfo{pages}{1499--1503}.
\newblock


\bibitem[Zhong et~al\mbox{.}(2020)]%
        {perturb}
\bibfield{author}{\bibinfo{person}{Haoti Zhong}, \bibinfo{person}{Cong Liao}, \bibinfo{person}{Anna~Cinzia Squicciarini}, \bibinfo{person}{Sencun Zhu}, {and} \bibinfo{person}{David~J. Miller}.} \bibinfo{year}{2020}\natexlab{}.
\newblock \showarticletitle{Backdoor Embedding in Convolutional Neural Network Models via Invisible Perturbation}. In \bibinfo{booktitle}{\emph{{CODASPY} '20: Tenth {ACM} Conference on Data and Application Security and Privacy, New Orleans, LA, USA, March 16-18, 2020}}. \bibinfo{publisher}{{ACM}}, \bibinfo{pages}{97--108}.
\newblock


\bibitem[zllrunning(2019)]%
        {faceparsing}
\bibfield{author}{\bibinfo{person}{zllrunning}.} \bibinfo{year}{2019}\natexlab{}.
\newblock \bibinfo{title}{face-parsing.PyTorch}.
\newblock \bibinfo{howpublished}{\url{https://github.com/zllrunning/face-parsing.PyTorch}}.
\newblock
\newblock
\shownote{Using modified BiSeNet for face parsing in PyTorch}.


\end{thebibliography}

% \appendix

\end{document}